\documentclass[10pt,twocolumn,letterpaper]{article}

\usepackage[pagenumbers]{cvpr} %

\newcommand{\showorhide}[1]{}

\newcommand{\rk}[1]{\showorhide{\textcolor{orange}{\textbf{rk}: #1}}} %

\definecolor{cvprblue}{rgb}{0.21,0.49,0.74}
\usepackage[pagebackref,breaklinks,colorlinks,allcolors=cvprblue]{hyperref}
\usepackage{amsmath}
\usepackage{amssymb}
\usepackage{multirow}
\usepackage{booktabs} %
\usepackage{algorithm}
\usepackage{algpseudocode}
\usepackage{tikz} %
\usetikzlibrary{positioning, arrows.meta} %

\def\paperID{} %
\def\confName{CVPR}
\def\confYear{2026}

\title{A Mixed Diet Makes DINO An Omnivorous Vision Encoder}

\author{
Rishabh Kabra$^{1,2}$ \quad
Maks Ovsjanikov$^{1}$ \quad
Drew A. Hudson$^{1}$ \quad
Ye Xia$^{1}$ \quad
Skanda Koppula$^{1,2}$ \quad \\
Andre Araujo$^{1}$ \quad
Joao Carreira$^{1}$ \quad
Niloy J. Mitra$^{2}$\\[0.5em]
$^{1}$Google DeepMind \qquad
$^{2}$University College London\\
{\tt\small \{rkabra,movsani,dorarad,yexia,skandak,andrearaujo,joaoluis\}@google.com \quad n.mitra@ucl.ac.uk}
}

\begin{document}
\maketitle

\begin{abstract}
Pre-trained vision encoders like DINOv2 have demonstrated exceptional performance on unimodal tasks. However, we observe that their features are poorly aligned across different visual modalities. For instance, the feature embedding for an RGB image and its corresponding depth map of the same scene exhibit a cosine similarity that is nearly identical to that of two random, unrelated images. To address this, we propose the Omnivorous Vision Encoder, a post-training framework that learns a modality-agnostic feature space. We fine-tune the encoder with a dual objective: first, to maximize the feature alignment between different modalities of the same scene; and second, a distillation objective that anchors the learned representations to a fully frozen teacher. The resulting student encoder becomes ``omnivorous'' by producing more consistent embeddings for a given scene, regardless of the input modality (RGB, Depth, Segmentation, etc.). This approach enables robust cross-modal understanding while retaining the discriminative semantics of the original foundation model. Omnivorous model weights are available at \url{https://github.com/google-deepmind/representations4d}.
\end{abstract}    
\section{Introduction}
\label{sec:intro}

Human perception exhibits remarkable stability: whether we view a scene in daylight, shadow, or through glasses, our internal representation of the scene remains largely invariant \cite{land1977retinex,zeki1983colour}. Ideally, a computer vision foundation model should possess this same ``omnivorous'' quality---mapping different modal views of the same scene (RGB, Depth, Segmentation) to almost identical points in its feature space. 

Our empirical analysis, however, reveals that popular off-the-shelf encoders fall short of this. We find that for leading models such as DINOv2 \cite{oquab2023dinov2}, feature maps for paired RGB, Depth, and Segmentation images are not well-aligned. Specifically, the cosine similarity between the features of an RGB image ($x_r$) and its corresponding depth map ($x_d$) is surprisingly low, often comparable to the similarity between unrelated scenes:
$ \cos(f(x_r), f(x_d)) \approx \cos(f(x_{r,1}), f(x_{r,2}))$, where $f$ is the pretrained encoder.

\begin{figure}[t]
    \centering
    \includegraphics[width=\linewidth]{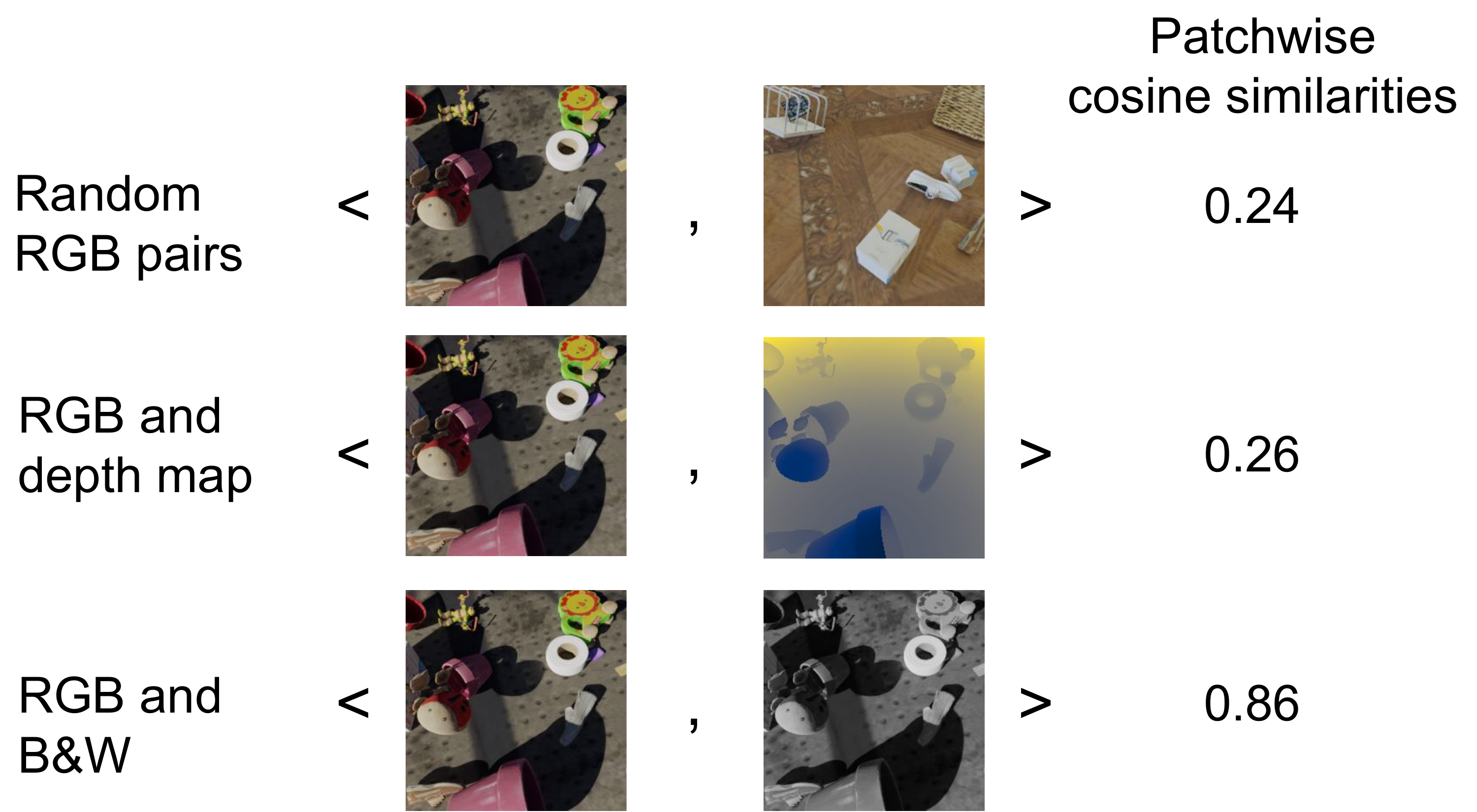}
    \caption{\textbf{Off-the-shelf vision encoders like DINO show poor cross-modal alignment.} We show the similarity in feature space between randomly paired RGB images (top), between RGB images and depth maps of the same scene (middle), and between RGB and grayscale images of the same scene (bottom). While the numbers vary depending on the dataset, the pattern of misalignment between visual modalities remains consistent. Our proposed adapter aligns these modalities in an existing feature space. \rk{Consider replacing for ScanNet images.}\vspace{-2mm}}
    \label{fig:motivation}
\end{figure}

We draw inspiration from the evolution of Natural Language Processing. Early NLP systems were language-specific \cite{sutskever2014sequence}. Later, it was demonstrated that aligning representations across languages \cite{johnson2017google,DBLP:journals/corr/abs-1710-11041}, or training shared multilingual encoders \cite{team2023gemini,NEURIPS2019_c04c19c2}, significantly improved generalization, particularly for low-resource languages. We argue that vision models face a similar inflection point (see Figure~\ref{fig:motivation}). By aligning abundant modalities (RGB) with structure-rich but scarcer signals (depth, segmentation), we can create a more robust, shared visual language.

Constructing this shared space presents a challenge. A trivial solution could simply collapse the feature space to achieve alignment, destroying the discriminative power of the encoder. Established methods like Contrastive Multiview Coding~(CMC) \cite{10.1007/978-3-030-58621-8_45} prevent collapse by pushing features apart when they are from different scenes. CMC relies on large sets of ``negative'' examples collected across datasets to ensure sample diversity, but these are typically limited to particular modalities (such as RGB) that are over-represented.

To align extremely unbalanced modalities, while ensuring strong discriminative power,  we propose a recipe that \textit{distills} cross-modal alignment into an existing foundation model. Our approach preserves the encoder's rich pre-trained priors, reduces the need to collect negative examples across sparse modalities, and is lightweight. We adopt a parameter-efficient teacher-student framework, with the ``student'' encoder initialized from the pretrained foundation model, and update only the final high-level processing blocks to align representations across visual modalities. We also introduce an anchoring loss to preserve the expressivity of the original feature space.

We couple this architectural recipe with two data-centric contributions designed to further discourage trivial alignment solutions. As context, modalities such as depth and segmentation can be represented in many different ways as images, for example through colormap choices. First, we observe that standard colormaps (e.g., grayscale or jet colormaps) allow models to shortcut alignment by relying on low-level channel statistics. To counter this, we colorize depth and segmentation maps using a natural color palette derived from the corresponding RGB image. This creates ``hard positives,'' making the contrastive task as hard as possible by forcing the network to align features based on structural content rather than superficial signals such as color histograms. Second, we introduce a modality blending strategy. Rather than treating modalities as discrete states, we randomly blend RGB, depth, and segmentation images during training. This encourages the student to learn a degree of invariance across a continuous space of modalities, resulting in an ``omnivorous'' encoder that remains robust even when visual inputs are ambiguous.

\section{Related Work}
\label{sec:related_work}

\paragraph{Unified encoders across visual modalities.}
A line of research seeks a single backbone that natively handles multiple \emph{visual} modalities. Omnivore \cite{Girdhar_2022_CVPR} trains one ViT to classify images, videos, and single-view 3D (RGB/depth-like inputs) with shared parameters, reporting benefits from joint training across modalities and an “omnivorous” design that reduces modality-specific heads.  Beyond purely visual streams, ImageBind \cite{Girdhar_2023_CVPR} learns a joint embedding that binds six modalities---image, text, audio, depth, thermal, and IMU---using only image-paired data, showing emergent alignment across unpaired modalities. Generalist architectures such as Uni-Perceiver \cite{Zhu_2022_CVPR,Li_2023_CVPR} unify many vision and language tasks with a single encoder–decoder interface, while Perceiver \cite{pmlr-v139-jaegle21a} and Perceiver IO \cite{Jaegle2022PerceiverIO} offer latent-bottlenecked Transformers designed to ingest heterogeneous inputs and emit structured outputs without modality-specific components. Autoregressive “all-in-one” systems like Unified-IO \cite{Lu2023UnifiedIO,Lu2024UnifiedIO2} extend to diverse modalities (RGB, depth, segmentation masks, language), demonstrating broad task coverage under a common tokenization of inputs and outputs. These works motivate learning a shared space, but they typically \emph{co-train} the backbone. By contrast, our approach targets alignment by fine-tuning a few layers on top of a \emph{frozen} unimodal backbone.

\paragraph{Aligning RGB, depth, and 3D representations.}
Numerous papers study RGB–depth (and 2D–3D) alignment during pretraining. CLIP2Point \cite{Huang2023CLIP2Point} transfers CLIP knowledge to 3D by \emph{image–depth} contrastive pretraining, providing a template for cross-modal InfoNCE on paired RGB/depth renders.  CoMAE \cite{Yang2023CoMAE} proposes a single-model hybrid scheme that first learns cross-modal alignment contrastively and then injects masked-autoencoding objectives, explicitly targeting RGB–depth representation sharing on SUN RGB-D \cite{song2015sun} and NYUv2 \cite{Silberman:ECCV12}. Mask3D \cite{Hou2023Mask3D} uses masked RGB-D pretraining to reconstruct depth and thereby embed 3D priors into a 2D backbone, an auxiliary signal that improves geometry awareness without labels. From a diagnostic perspective, \citet{Li2022CloserLook3D} provides a unified framework comparing perspective-, modality-, and format-invariance, and empirically studies which cross-format pairs matter most. More recent works explore progressive multimodal pretraining (e.g., contrastive then masked-autoencoding) \cite{Jamal_2025_CVPR} and spatial-aware \cite{CHEN2025103362} multi-scale contrastive losses for RGB-D dense prediction, reinforcing the value of explicit cross-modal objectives.

\paragraph{Adapters and parameter-efficient alignment.}
Rather than retraining large backbones, adapter methods add small trainable modules. ViT-Adapter \cite{Chen2023ViTAdapter} injects task/structure priors for dense prediction while keeping the ViT largely frozen, offering a strong blueprint for projector-style modules. For explicit cross-modal alignment with frozen encoders, MA-AVT \cite{Mahmud2024MAAVT} introduces blockwise contrastive alignment across audio-visual tokens in a parameter-efficient manner; its mechanics (blockwise objectives, shared/frozen trunk) inform projector designs that align modalities post-hoc.  Recent “modality-disentangle adapters” \cite{Zheng2024UnifiedRep} separate modality-invariant from modality-specific components—useful when wanting both a unified embedding and optional modality-specific residuals. %

\paragraph{Cross-modal distillation and source-free transfer.}
When some modalities are absent at test time, cross-modal knowledge distillation (CMKD) transfers supervision between modalities. SOCKET \cite{Ahmed2022SOCKET} performs source-free cross-modal transfer (e.g., RGB $\to$ depth/IR) without access to task-relevant source data, bridging modality gaps via paired task-irrelevant data and BN statistic matching.  Newer CMKD variants \cite{ferrod2025revisitingcrossmodalknowledgedistillation} for RGB-D semantic segmentation incorporate disentanglement and contrastive terms to structure the internal spaces of single-modality students, offering alternative formulations to projector-based alignment. These methods underscore the value of  contrastive/consistency losses across modalities and provide useful evaluation protocols (e.g., RGB-only inference after multimodal training).

\paragraph{Our contribution.}
Compared to unified co-training (Omnivore, ImageBind, Unified-IO) and RGB-D pretraining schemes (CLIP2Point, CoMAE, Mask3D), our method targets a pragmatic regime: \emph{post-hoc} alignment of heterogeneous modalities by learning a single lightweight projector $g$ on top of a fixed foundational backbone $f^*$. We use a loss that directly maximizes cross-modal agreement while preserving scene-level discrimination. This design maintains the deployment benefits of strong unimodal encoders (e.g., DINOv2), delivering an ``omnivorous'' embedding at inference time without full-model finetuning.
Our use of paired $(x^{M_1},x^{M_2})$ from the same scene and cosine/contrastive objectives follows established multimodal contrastive practice; TupleInfoNCE \cite{liu2021contrastive} also motivates constructing “hard” negatives by composing mismatched tuples.

\section{Method}
\label{sec:method}
Our goal is to learn a unified mapping from arbitrary visual modalities to a shared embedding space. We aim to achieve not only modality-invariant representations, but also a modality-agnostic encoder that utilizes a single set of shared parameters for all inputs.

\subsection{Architecture}
\begin{figure}
    \centering
    \includegraphics[width=\linewidth]{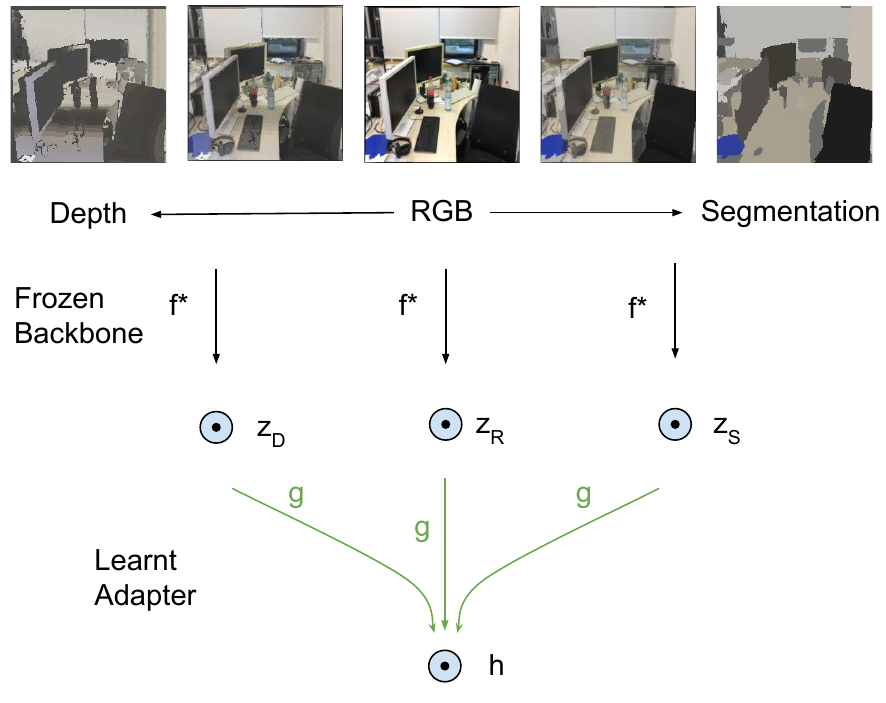}
    \caption{
    \textbf{Omnivorous Vision Encoder architecture. }
    A frozen encoder $f^*$ extracts features $z_m = f^*(x_m)$ from a spectrum of modalities denoted $m$ (Segmentation, RGB, Depth). A trainable modality-agnostic adapter $g$ maps these features into a common, aligned embedding space, producing a modality-invariant representation $h = g(z_m)$. A convenient implementation of this architecture uses the early layers of a pretrained network as the frozen part $f^*$, and the later layers as the adapter $g$.\vspace{-2mm}}
    \label{fig:model}
\end{figure}

We adopt a parameter-efficient teacher-student framework. We initialize a ``student'' encoder from the pre-trained foundation model. To balance stability and plasticity, the student shares the vast majority of its layers (the frozen backbone $f^*$) with the teacher, updating only the final high-level processing blocks (the head $g$). The teacher's head ($g^*$) remains frozen to serve as a stable anchor. 
By distilling knowledge from the teacher ($f_T = g^* \circ f^*$) into the student ($f_S = g \circ f^*$) while simultaneously maximizing cross-modal alignment, we prevent catastrophic forgetting. 
We will refer to $g$ as the ``adapter'' module to disambiguate it from task-specific ``heads'' trained later in our experiments. The architecture is depicted in Figure \ref{fig:model}. %

Let a scene be represented by a set of multimodal images $\{x_m | m \in M\}$, where $M$ is the set of modalities (e.g., $M = \{\text{RGB}, \text{Depth}, \text{Seg}\}$). 
For every input $x_m$, we compute two representations:
\begin{enumerate}[(i)]
    \item Teacher Output: $h^*_m = f_T(x_m) = g^*(f^*(x_m))$. This is the stable, pre-trained representation whose properties we aim to inherit.
    \item Student Output: $h_m = f_S(x_m) = g(f^*(x_m))$. This is the adapted representation we aim to align across modalities.
\end{enumerate}
Both $h^*_m$ and $h_m$ are $L_2$ normalized. Since our implementation distills from DINOv2, the network is a Vision Transformer \cite{dosovitskiy2020image}, comprising 12 blocks in the Base model. Unless mentioned otherwise, we freeze the first $L=8$ blocks and fine-tune the subsequent 4 for the student model.

\subsection{Data}
\label{sec:method_data}

Our data pipeline consists of three processing steps:

\noindent\textbf{1. Photometric augmentation (training).} We first apply standard brightness, contrast, hue, and saturation augmentations to the RGB image of a scene.

\noindent\textbf{2. Colorization (training and eval).} For a given (photometrically augmented) RGB image $x^{aug}_r$, we quantize its pixel values into 64 bins. These can then be used to colorize the corresponding segmentation or depth map, so the colorized maps $x_s$ and $x_d$ resemble the RGB image.

\noindent\textbf{3. Modality mixup (training) \cite{DBLP:journals/corr/abs-1710-09412}.} We derive an augmented segmentation image $x^{mixup}_s := (1-\alpha_s) x_s + \alpha_s x^{aug}_r$, and augmented depth image $x^{mixup}_d := (1-\alpha_d) x_d + \alpha_d x^{aug}_r$. The blending parameters $\alpha_s$ and $\alpha_d$ are stochastically sampled, independently of each other, per datapoint. 

Theoretically, the space of mixed-up segmentations $M_{s} := \{x^{mixup}_s | (x_s, x_r) \in \chi, \alpha_s \in [0, 1]\}$ and the space of mixed-up depth images $M_{d} := \{x^{mixup}_d | (x_d, x_r) \in \chi, \alpha_d \in [0, 1]\}$ together span a continuous space of modalities, loosely: Depth $\leftrightarrow$ RGB $\leftrightarrow$ Segmentation. In practice, we restrict the range of both $\alpha$'s to [0, 0.5] while training to prevent depth and segmentation images from looking too similar to the RGB image. We ablate the choice of $\alpha_{max}=0.5$ in Sec \ref{sec:ablations}. For evaluation (e.g., inter-modal retrieval), we set the $\alpha$'s to 0.

 Figure \ref{fig:training_data} illustrates our training data spanning six datasets (detailed further in Appendix \ref{app:training_data}).

\subsection{Loss}
\paragraph{Symmetric Cross-Modal Alignment}
To create a unified representation space, we employ a symmetric alignment strategy. We aim for the student embeddings from the same scene but different modalities to be close, while embeddings from different scenes should be distinct.

We use the InfoNCE (Information Noise-Contrastive Estimation) loss \cite{oord2018representation}. Given a batch of $N$ scenes, we define positive pairs as the student embeddings of two different modalities from the same scene $i$, $(h_{m_1}^{(i)}, h_{m_2}^{(i)})$, and negative pairs as embeddings from different scenes $(h_{m_1}^{(i)}, h_{m_2}^{(j)})$ where $i \neq j$. The loss for a specific pair of modalities $(m_1, m_2)$ is:
\begin{equation}
\begin{split}
\mathcal{L}_{\text{InfoNCE}}&(m_1, m_2) = \\
    &-\frac{1}{N} \sum_{i=1}^{N} \log \frac{\exp_{\tau}(\text{sim}(h_{m_1}^{(i)}, h_{m_2}^{(i)}))}{\sum_{j=1}^{N} \exp_{\tau}(\text{sim}(h_{m_1}^{(i)}, h_{m_2}^{(j)}))}
\end{split}
\end{equation}

Here $\text{sim}(\cdot, \cdot)$ denotes the cosine similarity, $\exp_{\tau}(x) = \exp(x/\tau)$, and $\tau$ is a learned temperature parameter (clipped to [0., 100.]). The total alignment loss, $\mathcal{L}_{\text{align}}$, is the average of the symmetric InfoNCE losses computed over all modality pairs in the adapted space, i.e.:
\begin{align}
\mathcal{L}_{\text{align}}    = \frac{1}{3}
\sum^3_{k_1=1} \sum^3_{k_2>k_1}\mathcal{L}_\text{InfoNCE}(m_{k_1}, m_{k_2}) 
\end{align} 
The three choices of pairs of modalities $(m_{k_1}, m_{k_2})$ lead to the following pairs of augmented features: $(h^{aug}_r, h^{mixup}_s),(h^{mixup}_s, h^{mixup}_d),$ and $(h^{mixup}_d, h^{aug}_r).$ This symmetric approach avoids the conflicting optimization targets inherent in aligning adapted features to potentially misaligned frozen features.

\paragraph{Anchoring Loss}
While $\mathcal{L}_{\text{align}}$ brings modalities together, it can lead to ``representational drift'' or collapse. The adapter might learn a trivial solution that satisfies alignment but discards the rich semantic information captured by the frozen backbone $f^*$.
To mitigate this, we introduce an anchoring loss, $\mathcal{L}_{\text{anchor}}$. This loss acts as a distillation mechanism, encouraging the student's output $h_m$ to remain close to the teacher's output $h^*_m$ of the \textit{same} modality. We use the cosine distance for this objective:
\begin{align} \mathcal{L}_{\text{anchor}} = \frac{1}{|M|} \sum_{m \in M} (1 - \text{sim}(h_m, h^*_m))
\end{align}

By anchoring $h_m$ to the stable, pre-trained space of $h^*_m$, we preserve the discriminative power of the original representation.

\paragraph
{Total Objective and Implementation.}
\label{sec:total_objective}
The final training objective is a weighted sum of the two losses:
\begin{align}
\mathcal{L}_{\text{total}} = \mathcal{L}_{\text{align}} + \lambda_{\text{anchor}} \mathcal{L}_{\text{anchor}}. 
\end{align}

The hyperparameter $\lambda_{\text{anchor}}$ balances the trade-off between achieving cross-modal alignment and preserving the semantics of the input modality. A higher $\lambda_{\text{anchor}}$ emphasizes fidelity to the teacher's semantics, while a lower value prioritizes alignment. A non-zero $\lambda_{\text{anchor}}$ is crucial when using symmetric alignment to prevent degenerate solutions. We use a default value of $\lambda_{anchor}=10$.

We compute the losses separately for the class token and the dense tokens output by the network. In the latter case, we subsample 64 dense tokens for each image before computing the loss. We use a mask to ensure we do not use intra-image dense tokens as negative examples for $\mathcal{L}_{\text{InfoNCE}}$.

\begin{figure}
    \centering
    \includegraphics[width=\linewidth]{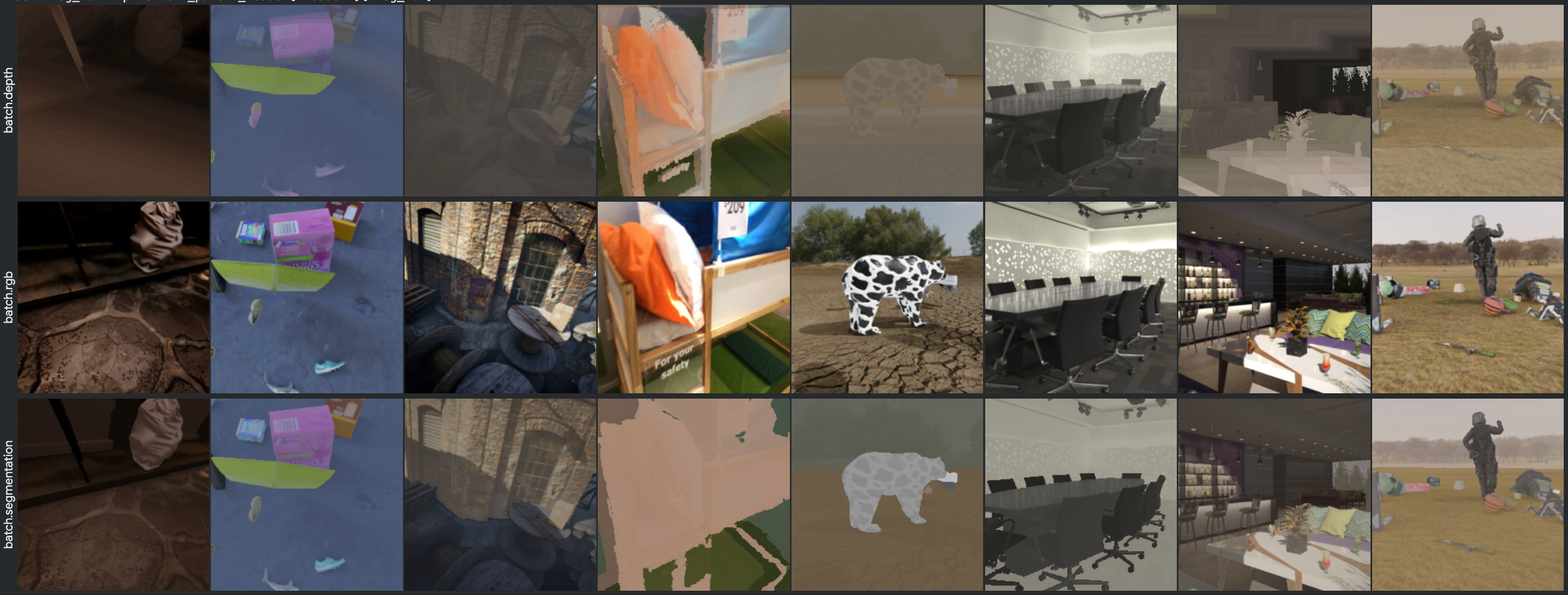}
    \caption{\textbf{Training data:} depth and segmentation maps are first colorized using a \emph{natural} color palette derived from the corresponding RGB image. We then apply a data augmentation: we blend the colorized depth image with up to 50\% of the RGB image (and likewise for the segmentation image). The compositing alpha is randomly sampled (between 0\% to 50\%) for each datapoint. The idea is to interpolate between the modalities (Depth $\leftrightarrow$ RGB $\leftrightarrow$ Seg) smoothly and teach the model a degree of invariance across the full spectrum, while also providing more negative examples for between-scene contrastive learning. Other potential benefits: the augmentation (i) makes our representations naturally invariant to scene lighting; and (ii) helps us cope with imperfect depth and segmentation values.\vspace{-2mm}
    }
    \label{fig:training_data}
\end{figure}

\section{Experiments}
\label{sec:experiments}

We evaluate a single Omnivorous checkpoint versus DINOv2 across the following settings: in Section \ref{sec:intermodal_retrieval} we assess retrieval between modalities without any additional training. In Section \ref{sec:downstream_tasks}, we train linear and non-linear heads to evaluate on downstream tasks (classification, monocular depth prediction, and segmentation) on novel datasets. In Section \ref{sec:crossmodal_transfer}, we train a depth prediction head from RGB images, then switch up the input modality beyond the training distribution. Finally, in Section \ref{sec:ablations} we present a set of ablations for our training pipeline.

\subsection{Inter-Modal Retrieval}
\label{sec:intermodal_retrieval}

To assess the alignment of our features, we perform cross-modal retrieval evaluations. This task measures the ability to retrieve the correct scene in a target modality (e.g., Depth) given a query in a source modality (e.g., RGB).

\vspace{1mm}
\noindent\textbf{Evaluation Protocol.}
We extract features for all scenes in the test sets of MOVi \cite{greff2022kubric}, ScanNet \cite{dai2017scannet}, and TartanAir \cite{wang2020tartanair}. We consider three modalities: RGB, Depth, and Segmentation. For every scene, we extract features using both the standard [CLS] token and Global Average Pooling (GAP) of the dense feature map. All feature vectors are $L_2$ normalized. We compute the pairwise cosine similarity between the query and gallery sets.
We report standard information retrieval metrics: Recall at $k$ ($R@k$ for $k=\{1, 5\}$), Mean Average Precision (mAP), and Median Rank (MedR).

To capture the holism of the shared space, the results reported in Table~\ref{tab:cross_modal_retrieval} are averaged over all 6 unique directed modality pairs (RGB$\rightarrow$Depth, Depth$\rightarrow$RGB, RGB$\rightarrow$Seg, Seg$\rightarrow$RGB, Depth$\rightarrow$Seg, Seg$\rightarrow$Depth).

\vspace{1mm}
\noindent\textbf{Results.}
Table~\ref{tab:cross_modal_retrieval} compares our Omnivorous encoder against the frozen DINOv2 baseline.
The baseline exhibits significant misalignment across modalities. On ScanNet, the DINOv2 features yield a Median Rank of 401.8 (GAP) and 382.5 (TOK), indicating that the embeddings for different views of the same scene are far apart in the latent space.

In contrast, the Omnivorous adapter significantly improves alignment without requiring fine-tuning of the backbone. On ScanNet (GAP), our method improves $R@1$ from 4.6\% to 46.1\% and reduces the Median Rank to 2.0.
On the synthetic datasets (MOVi and TartanAir), where domain gaps are smaller, the alignment is near-perfect. For example, on MOVi, we achieve an $R@1$ of 86.2\% compared to the baseline's 15.5\%.

\begin{table*}[]
    \centering
    \caption{\textbf{Cross-modal retrieval: average results across all 6 directed modality pairs.} We sample 1 frame per test video, yielding $N$ queries and targets per dataset.
    GAP: Global Average Pooling of dense features. TOK: CLS Token embedding. \rk{Add N per dataset to make sense of MedR.}}
    \label{tab:cross_modal_retrieval}
\begin{tabular}{lllcccc}
\toprule
dataset & feature type & model & R@1 $\uparrow$ & R@5 $\uparrow$ & mAP $\uparrow$ & MedR $\downarrow$ \\
\midrule
movi & \multirow[t]{2}{*}{gap} & DINOv2 ViT-B/14 & 15.5 & 33.1 & 25.2 & 19.3 \\
\multirow[t]{3}{*}{(N=128)} &  & Omnivorous ViT-B/14 & \textbf{86.2} & \textbf{96.5} & \textbf{90.9} & \textbf{1.0} \\
\cline{2-7}
 & \multirow[t]{2}{*}{tok} & DINOv2 ViT-B/14 & 18.2 & 34.5 & 27.2 & 16.8 \\
 &  & Omnivorous ViT-B/14 & \textbf{76.6} & \textbf{92.7} & \textbf{83.4} & \textbf{1.0} \\
\cline{1-7} \cline{2-7}
scannet & \multirow[t]{2}{*}{gap} & DINOv2 ViT-B/14 & 4.6 & 10.8 & 8.1 & 401.8 \\
\multirow[t]{4}{*}{(N=3072)} &  & Omnivorous ViT-B/14 & \textbf{46.1} & \textbf{71.4} & \textbf{57.7} & \textbf{2.0} \\
\cline{2-7}
 & \multirow[t]{2}{*}{tok} & DINOv2 ViT-B/14 & 3.9 & 9.0 & 6.9 & 382.5 \\
 &  & Omnivorous ViT-B/14 & \textbf{30.2} & \textbf{55.8} & \textbf{42.2} & \textbf{5.3} \\
\cline{1-7} \cline{2-7}
tartanair & \multirow[t]{2}{*}{gap} & DINOv2 ViT-B/14 & 46.6 & 68.5 & 57.1 & 1.8 \\
\multirow[t]{4}{*}{(N=128)} &  & Omnivorous ViT-B/14 & \textbf{90.6} & \textbf{99.2} & \textbf{94.6} & \textbf{1.0} \\
\cline{2-7}
 & \multirow[t]{2}{*}{tok} & DINOv2 ViT-B/14 & 43.4 & 66.7 & 54.7 & 2.1 \\
 &  & Omnivorous ViT-B/14 & \textbf{84.5} & \textbf{98.4} & \textbf{90.5} & \textbf{1.0} \\
\cline{1-7} \cline{2-7}
\bottomrule
\end{tabular}

\end{table*}

\begin{table*}[]
\centering
\caption{\textbf{Downstream evals: monocular depth prediction and segmentation.} We train either a Dense Prediction Transformer (DPT) or linear head on top of the frozen ViT backbone. \textbf{Depth:} The training minimizes a scale-invariant gradient loss and an edge-aware gradient loss. Evaluation is conducted on datasets like NYUv2 using standard metrics such as RMSE and threshold accuracy ($\delta_i = 1.25^i$). \textbf{Segmentation:} The decoder heads are trained for pixel-wise classification. During evaluation, we compute Mean Intersection-over-Union (mIoU) by aggregating confusion matrices across batches.
}
\label{tab:depth_and_seg}
\begin{tabular}{llccccccc}
\toprule
 &  & \multicolumn{2}{c}{depth delta1 $\uparrow$} & \multicolumn{2}{c}{depth rmse $\downarrow$} & \multicolumn{3}{c}{segmentation mean iou $\uparrow$} \\
 & dataset & navi probe3d & nyuv2 & navi probe3d & nyuv2 & ade20k & cityscapes & pascal voc \\
readout & model &  &  &  &  &  &  &  \\
\midrule
\multirow[t]{2}{*}{Linear} & DINOv2 ViT-B/14 & 0.697 & 0.875 & 0.076 & 0.405 & 0.463 & 0.622 & 0.814 \\
 & Omnivorous ViT-B/14 & \textbf{0.706} & \textbf{0.896} & \textbf{0.074} & \textbf{0.377} & \textbf{0.475} & \textbf{0.632} & \textbf{0.826} \\
\cline{1-9}
\multirow[t]{2}{*}{DPT} & DINOv2 ViT-B/14 & 0.779 & 0.948 & 0.061 & 0.297 & 0.496 & \textbf{0.737} & 0.855 \\
 & Omnivorous ViT-B/14 & \textbf{0.781} & 0.948 & 0.061 & 0.297 & \textbf{0.505} & 0.732 & \textbf{0.857} \\
\cline{1-9}
\bottomrule
\end{tabular}
\vspace{-2mm}
\end{table*}

\begin{table}[]
\centering
\caption{\textbf{Downstream eval: linear-probe classification} on ImageNet. We sweep over five learning rates, picking the best one for each row. TOK: CLS Token embedding. TOK \& GAP: both the CLS embedding and Average-Pooled dense features are used. \rk{Should we report accuracy at worst learning rate too?}}
\label{tab:linear_probe}
\begin{tabular}{llc}
\toprule
feature type & model & accuracy $\uparrow$ \\
\midrule
\multirow[t]{4}{*}{tok} & DINOv2 ViT-B/14 & 0.801 \\
 & Omnivorous ViT-B/14 & \textbf{0.835} \\
\multirow[t]{4}{*}{tok \& gap} & DINOv2 ViT-B/14 & 0.804 \\
 & Omnivorous ViT-B/14 & \textbf{0.838} \\
\bottomrule
\end{tabular}
\end{table}

\begin{table*}[]
\centering
\caption{\textbf{Downstream eval: k-NN classification}. On ImageNet \cite{deng2009imagenet}, we follow the standard DINO evaluation protocol by using soft voting among the top-$k$ neighbors (weighted by similarity) to predict classes, sweeping over multiple $k$ values (e.g., 10, 20, 100) to report the best top-1 accuracy. On all other datasets (iNaturalist \cite{vanhorn2018inaturalistspeciesclassificationdetection}, SOP \cite{song2016deep}, GLDv2 \cite{weyand2020GLDv2}, RP2K \cite{peng2020rp2k}, Food2k \cite{min2023largescalevisualfood}), we use universal embeddings, evaluating the ``hard'' k-NN accuracy by matching test query embeddings against a training index of embeddings.\vspace{-1mm}}
\label{tab:knn}
\begin{tabular}{lrrrrrr}
\toprule
 model & imagenet soft & inat & sop & gldv2 & rp2k & food2k \\
\midrule
DINOv2 ViT-B/14 & 81.936 & \textbf{78.53} & 54.39 & \textbf{51.90} & 66.83 & 51.90 \\
Omnivorous ViT-B/14 & \textbf{81.974} & 77.49 & \textbf{54.69} & 50.13 & \textbf{70.48} & \textbf{52.14} \\
\bottomrule
\end{tabular}
\end{table*}

\subsection{Cross-Dataset and Cross-Task Transfer}
\label{sec:downstream_tasks}

We run a suite of downstream evaluations to assess whether the Omnivorous encoder successfully aligns modalities without compromising the semantic power of the underlying foundation model. Following the protocols established in DINOv2 and Probe3D~\cite{El_Banani_2024_CVPR}, we evaluate on monocular depth estimation, semantic segmentation, and classification. Further results on normals estimation and 3D correspondence are in Appendix \ref{app:extended_ablations}.

\vspace{1mm}
\noindent\textbf{Monocular Depth Estimation.}
We evaluate geometric awareness by training lightweight decoders on top of the now-frozen student network. We report results on NYUv2 \cite{Silberman:ECCV12} and NAVI \cite{jampani2023navi} in Table~\ref{tab:depth_and_seg}. When using a simple Linear readout, the Omnivorous encoder outperforms DINOv2, reducing the RMSE from 0.405 to 0.377, and improving the $\delta_1$ accuracy (percentage of correctly predicted depth pixels) from 0.875 to 0.896.
With the more expressive DPT decoder, performance remains at parity with the strong DINOv2 baseline (0.297 RMSE), confirming that our adapter preserves the fine-grained geometric information necessary for dense prediction.

\vspace{1mm}
\noindent\textbf{Semantic Segmentation.}
To verify the utility of our aligned representations for dense semantic tasks, we evaluate on ADE20k \cite{zhou2017scene}, Cityscapes \cite{cordts2016cityscapes}, and Pascal VOC \cite{Everingham10} (Table~\ref{tab:depth_and_seg}).
Our method achieves competitive performance, often surpassing the unimodal baseline. Notably, on ADE20k with a Linear readout, we improve the mIoU from 0.463 to 0.475. Similarly, on Cityscapes (Linear), we observe a gain from 0.622 to 0.632.
These results demonstrate that enforcing alignment between RGB, depth, and segmentation maps does not degrade the high-level semantic understanding required for segmentation tasks; in fact, the multimodal regularization appears to offer slight benefits in generalization.

\vspace{1mm}
\noindent\textbf{Classification.}
We first assess the linear separability of our representations by training a linear classifier on top of the frozen backbone for ImageNet-1k (Table \ref{tab:linear_probe}). Our Omnivorous encoder demonstrates a substantial improvement over DINOv2 (top-1 accuracy of 83.8\% compared to 80.4\%). This marked improvement suggests that aligning structural modalities (depth, segmentation) with RGB enriches the semantic density of the shared feature space, making it significantly more discriminative for standard classification.

We further examine k-Nearest Neighbor (k-NN) classification to ensure the anchoring loss $\mathcal{L}_{anchor}$ effectively mitigated representational drift (Table \ref{tab:knn}). On ImageNet (soft voting), k-NN performance remains effectively at parity with the teacher (81.97\% vs 81.94\%), confirming that our student encoder has not forgotten the original pre-training. The results on downstream transfer datasets are mixed: we observe notable gains on RP2K (+3.65\%), suggesting improved robustness for object-centric tasks. However, we note slight regressions on fine-grained datasets like iNaturalist and Google Landmarks v2. This could be explained by our training mix, which includes a significant amount of simulated multi-object data. We conclude that while ``omnivorous'' alignment generally preserves semantics, the choice of training data nevertheless matters.

\subsection{Zero-Shot Cross-Modal Transfer}
\label{sec:crossmodal_transfer}

A key promise of a unified feature space is the ability to train a task head on one modality and deploy it on another without retraining. To test this, we train a depth prediction head (Linear or DPT) on the NYUv2 dataset using only RGB images as input. We then evaluate this head on the PACE \cite{you2024pace} dataset, but we switch the input modality to Segmentation maps (which are within our Omnivorous backbone's training distribution) and NOCS maps (which are out-of-distribution for both backbones).

As shown in Table~\ref{tab:cross_modal_transfer}, the frozen DINOv2 baseline fails catastrophically when the modality is switched.
When fed Segmentation maps, the DINOv2 Linear head yields an RMSE of 1.536 (meters), which is effectively random guessing.
In contrast, the Omnivorous encoder—which has mapped Segmentation inputs to the same semantic space as the RGB training data—achieves an RMSE of 0.532.

This advantage extends to unseen modalities. When testing on NOCS (Normalized Object Coordinate Space) \cite{wang2019normalized}, which neither model saw during training, the Omnivorous encoder still significantly outperforms the baseline (RMSE 1.075 vs 1.996).
This suggests that by learning to align visual modalities, the Omnivorous encoder learns a more general representation that is more robust to modality shifts than the RGB-specific DINOv2 backbone.

\begin{table*}
    \centering
    \caption{\textbf{Cross-modal transfer on the depth prediction task}. Readout heads are trained on RGB images, but tested zero-shot on two novel modalities: Seg (within-distribution for Omnivorous) and NOCS images (out-of-distribution for both  Omnivorous and DINO). The heads are trained on NYUv2, evaluated on PACE. We also show qualitative results for both models.\vspace{-2mm}}
    \label{tab:cross_modal_transfer}
    \begin{minipage}[c]{0.55\linewidth}
        \centering
    \begin{tabular}{lllll}
    \toprule
    input & readout & model & delta1 $\uparrow$ & rmse $\downarrow$ \\
    \midrule
    \multirow[t]{4}{*}{rgb} & \multirow[t]{2}{*}{Linear} & DINOv2 ViT-B/14 & 0.108 & 0.842 \\
     &  & Omnivorous ViT-B/14 & \textbf{0.146} & \textbf{0.671} \\
    \cline{2-5}
     & \multirow[t]{2}{*}{DPT} & DINOv2 ViT-B/14 & 0.420 & 0.318 \\
     &  & Omnivorous ViT-B/14 & \textbf{0.463} & \textbf{0.290} \\
    \cline{1-5} \cline{2-5}
    \multirow[t]{4}{*}{seg} & \multirow[t]{2}{*}{Linear} & DINOv2 ViT-B/14 & 0.003 & 1.536 \\
     &  & Omnivorous ViT-B/14 & \textbf{0.184} & \textbf{0.532} \\
    \cline{2-5}
     & \multirow[t]{2}{*}{DPT} & DINOv2 ViT-B/14 & 0.042 & 0.792 \\
     &  & Omnivorous ViT-B/14 & \textbf{0.169} & \textbf{0.507} \\
    \cline{1-5} \cline{2-5}
    \multirow[t]{4}{*}{nocs} & \multirow[t]{2}{*}{Linear} & DINOv2 ViT-B/14 & 0.001 & 1.996 \\
     &  & Omnivorous ViT-B/14 & \textbf{0.023} & \textbf{1.075} \\
    \cline{2-5}
     & \multirow[t]{2}{*}{DPT} & DINOv2 ViT-B/14 & 0.014 & 0.979 \\
     &  & Omnivorous ViT-B/14 & \textbf{0.029} & \textbf{0.822} \\
    \cline{1-5} \cline{2-5}
    \bottomrule
    \end{tabular}
    \end{minipage}%
    \hfill %
    \begin{minipage}[c]{0.2\linewidth}
        \centering
        DINOv2 (DPT)
        \vspace{0cm}
        \includegraphics[width=\linewidth, height=2.2cm, keepaspectratio]{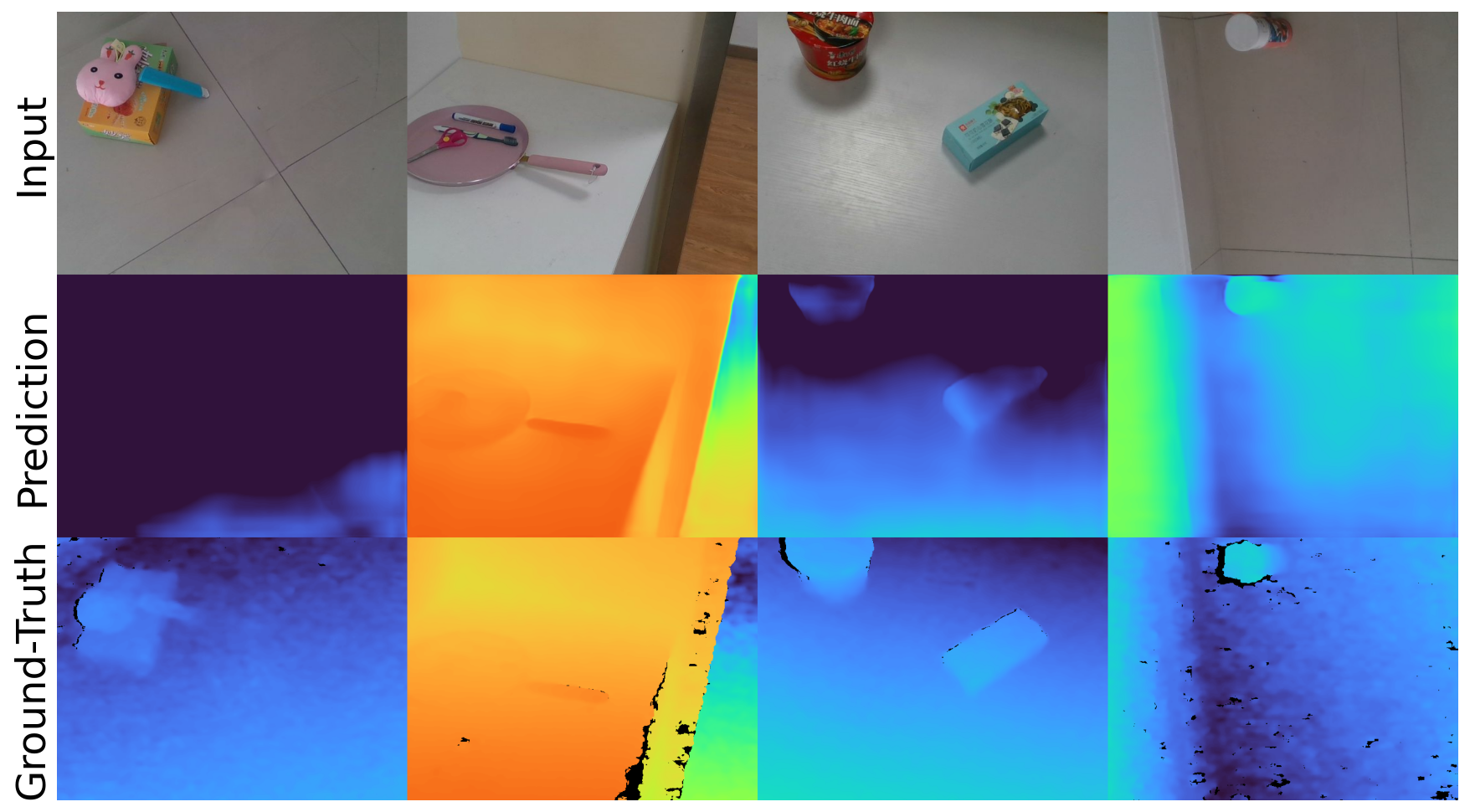}

        \includegraphics[width=\linewidth, height=2.2cm, keepaspectratio]{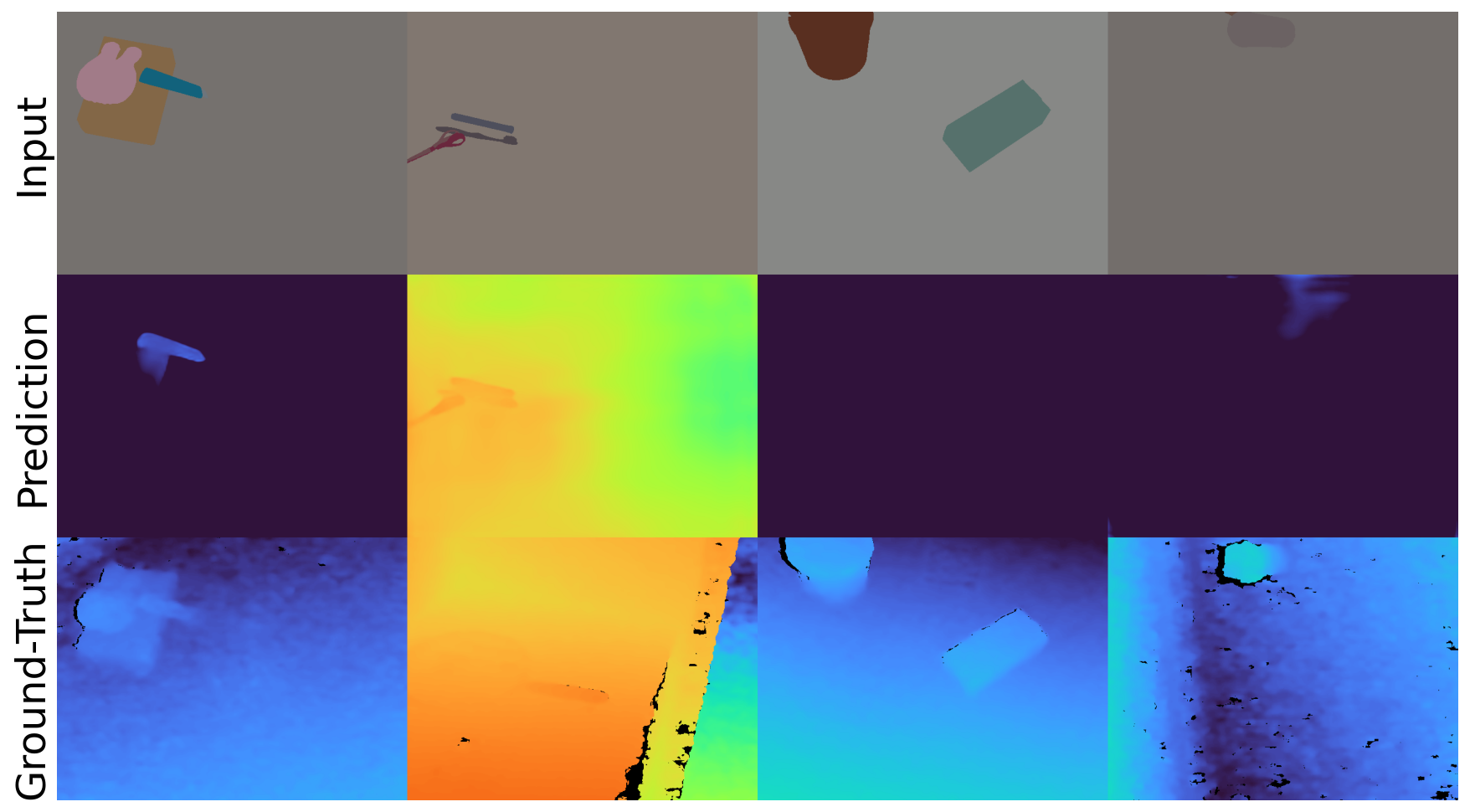}

        \includegraphics[width=\linewidth, height=2.2cm, keepaspectratio]{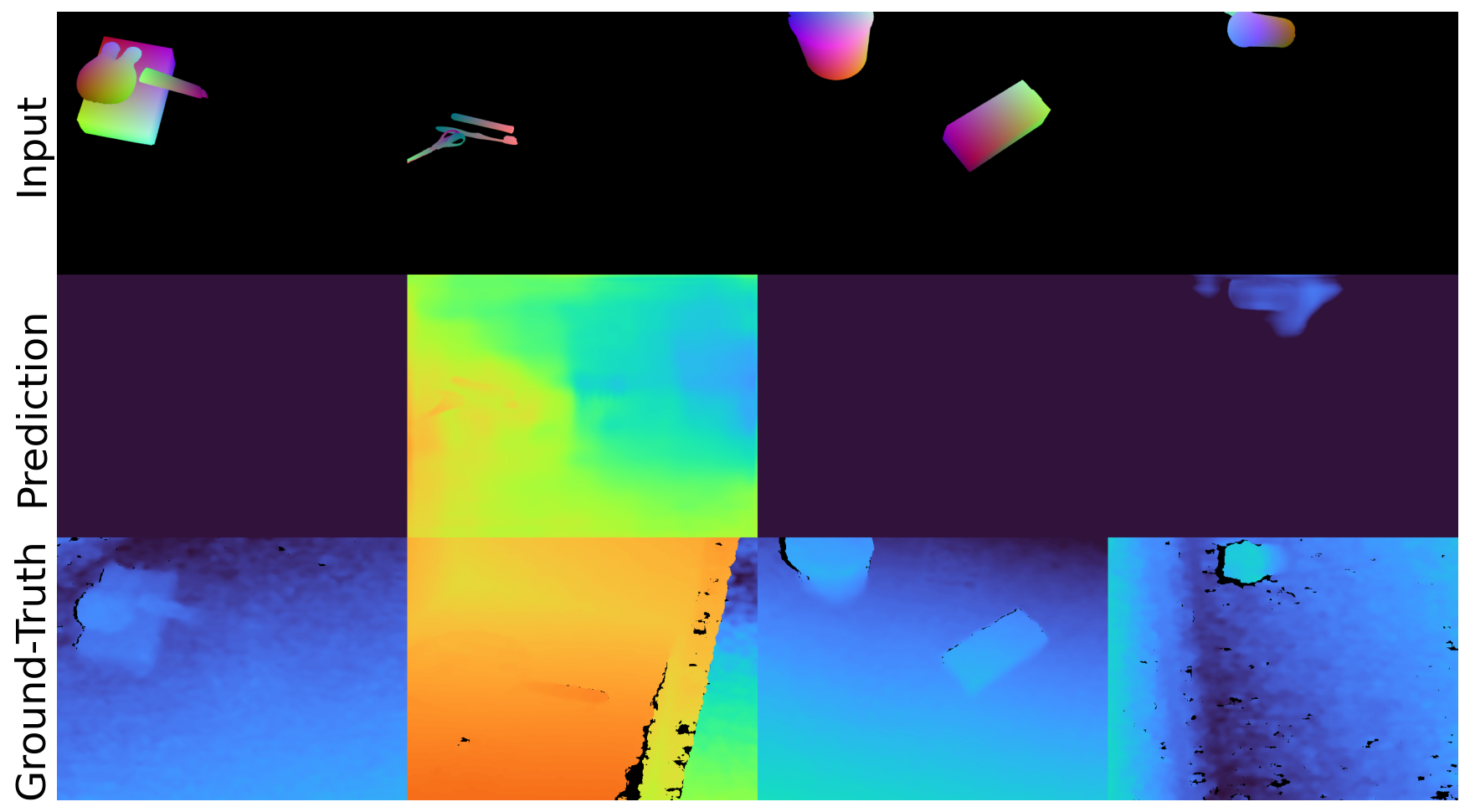}
    \end{minipage}
    \hfill %
    \begin{minipage}[c]{0.2\linewidth}
        \centering
        Omnivorous (DPT)
        \vspace{0cm}
        \includegraphics[width=\linewidth, height=2.2cm, keepaspectratio]{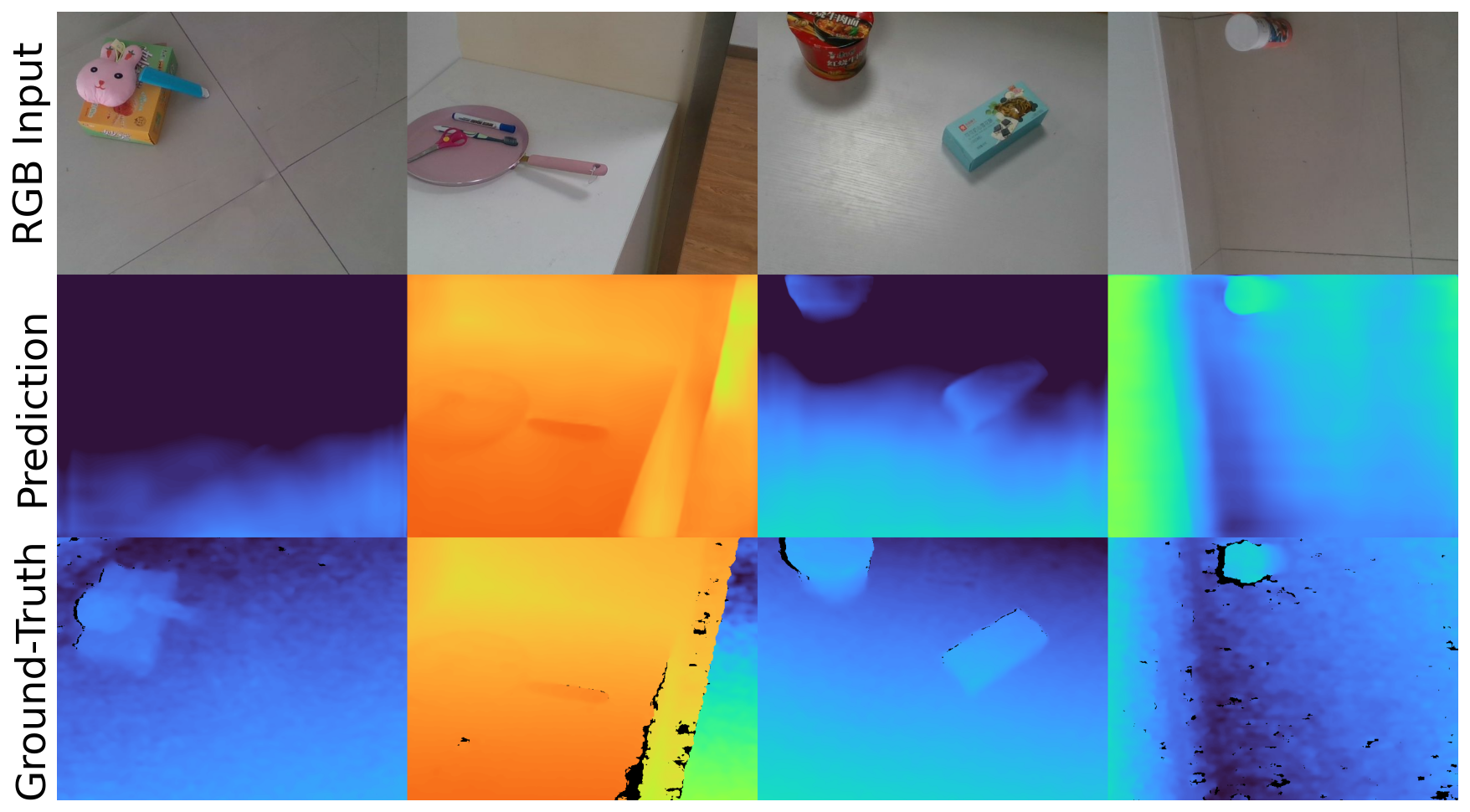}

        \includegraphics[width=\linewidth, height=2.2cm, keepaspectratio]{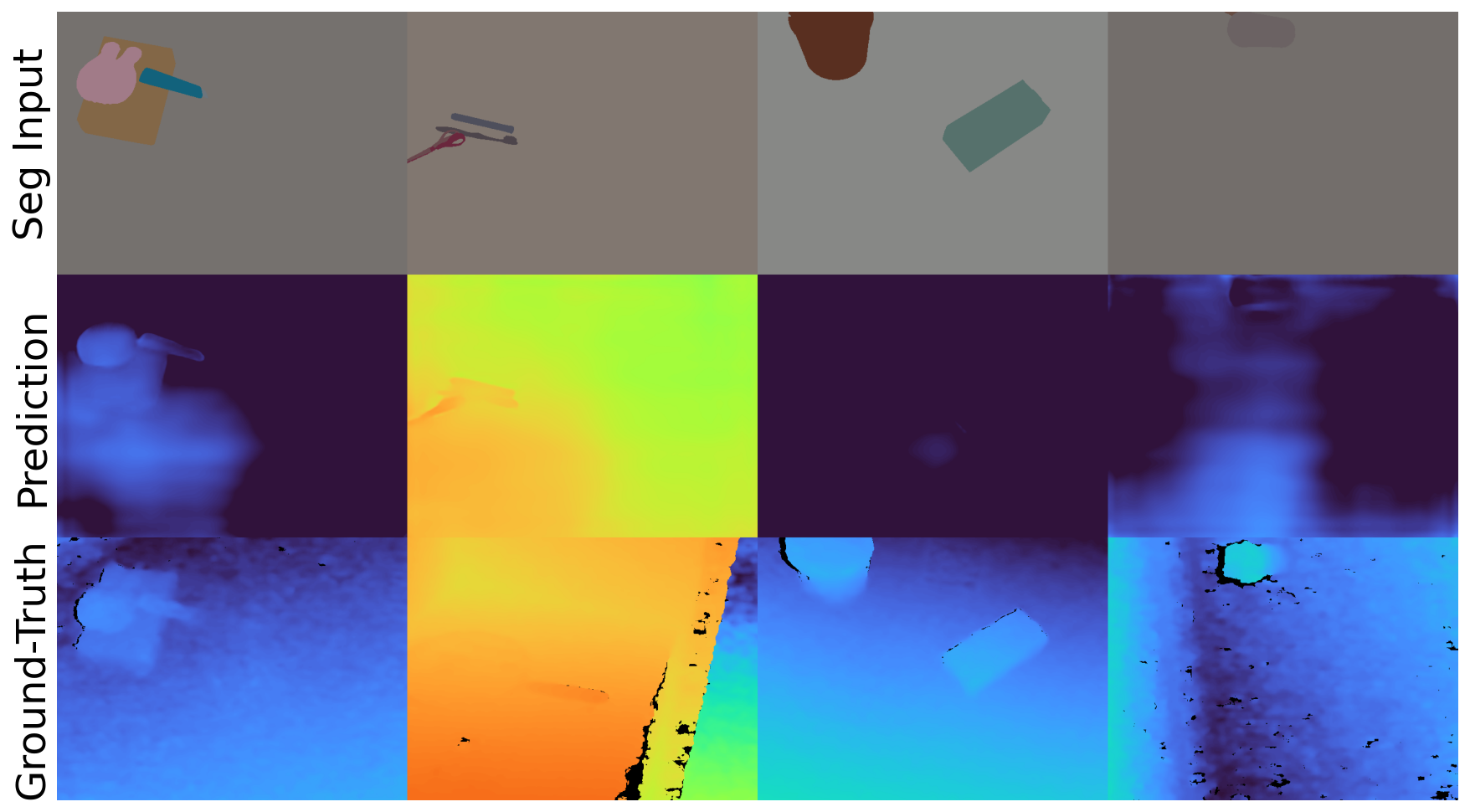}

        \includegraphics[width=\linewidth, height=2.2cm, keepaspectratio]{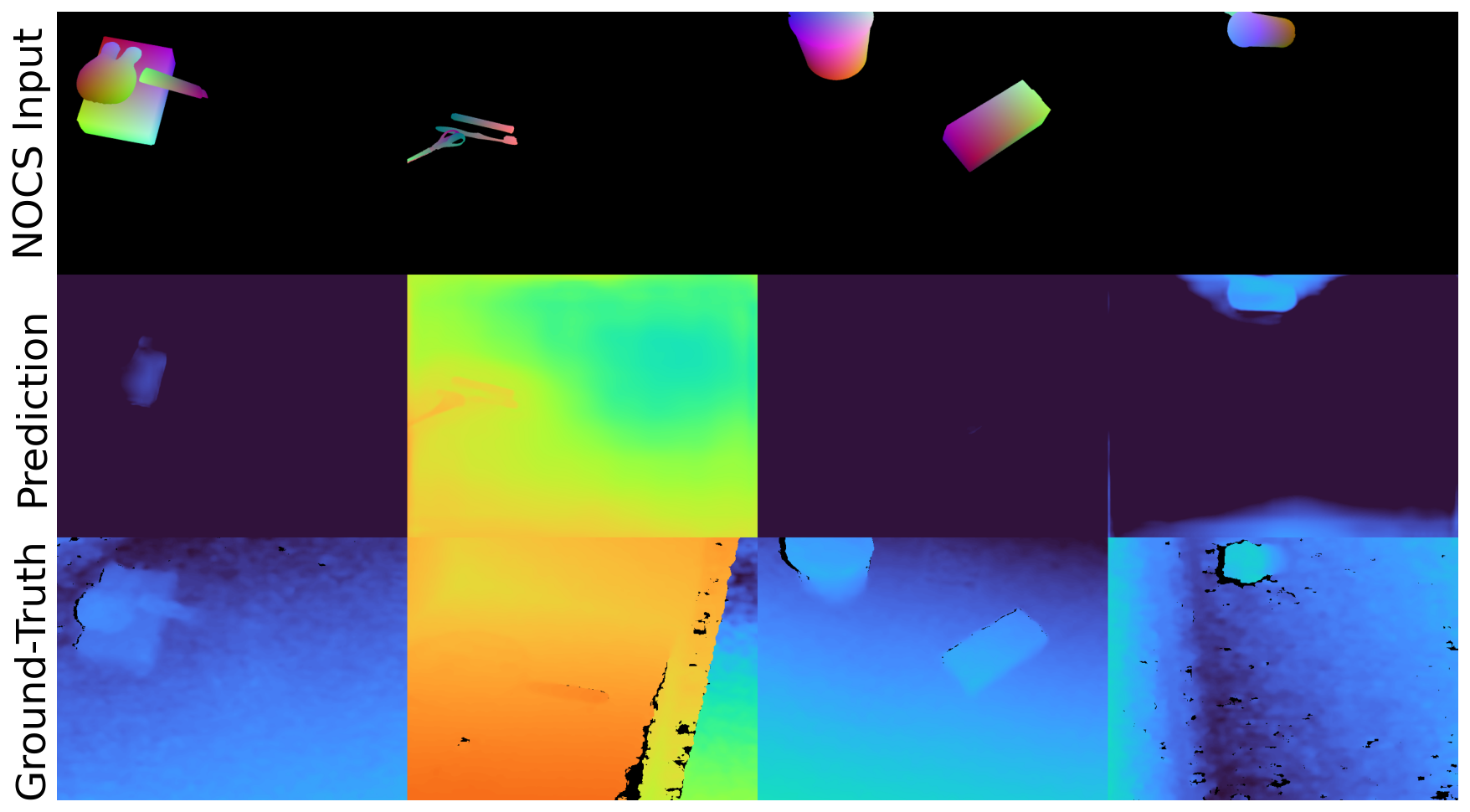}
    \end{minipage}
    \end{table*}

\subsection{Ablations}
\label{sec:ablations}

\paragraph{Loss.} A key component of our method is the hyperparameter $\lambda_{anchor}$, which balances the symmetric alignment loss ($\mathcal{L}_{align}$) and the anchoring loss ($\mathcal{L}_{anchor}$) as defined in Section \ref{sec:total_objective}. This parameter explicitly controls the trade-off between cross-modal alignment and preserving the original discriminative power of the frozen teacher's features.

\begin{figure}[htbp]
  \centering
  
  \begin{subfigure}[b]{\linewidth}
    \centering
    \includegraphics[width=\linewidth]{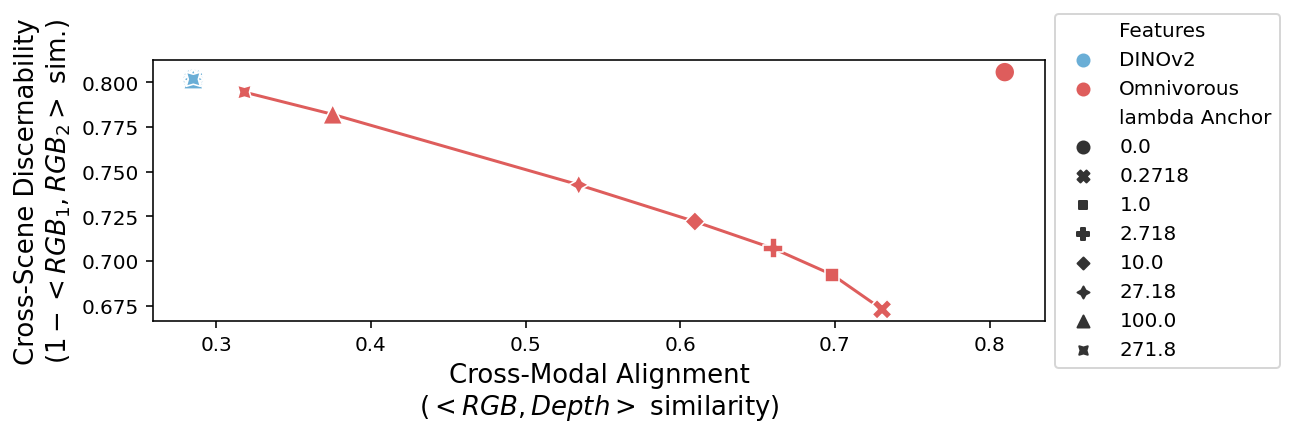}
    \caption{Performance frontier (Alignment vs. Discernibility)\vspace{-2mm}}
    \label{fig:frontier-plot}
  \end{subfigure}
  
  \vspace{1em} 

  \begin{subfigure}[b]{0.9\linewidth}
    \centering
    \includegraphics[width=\linewidth]{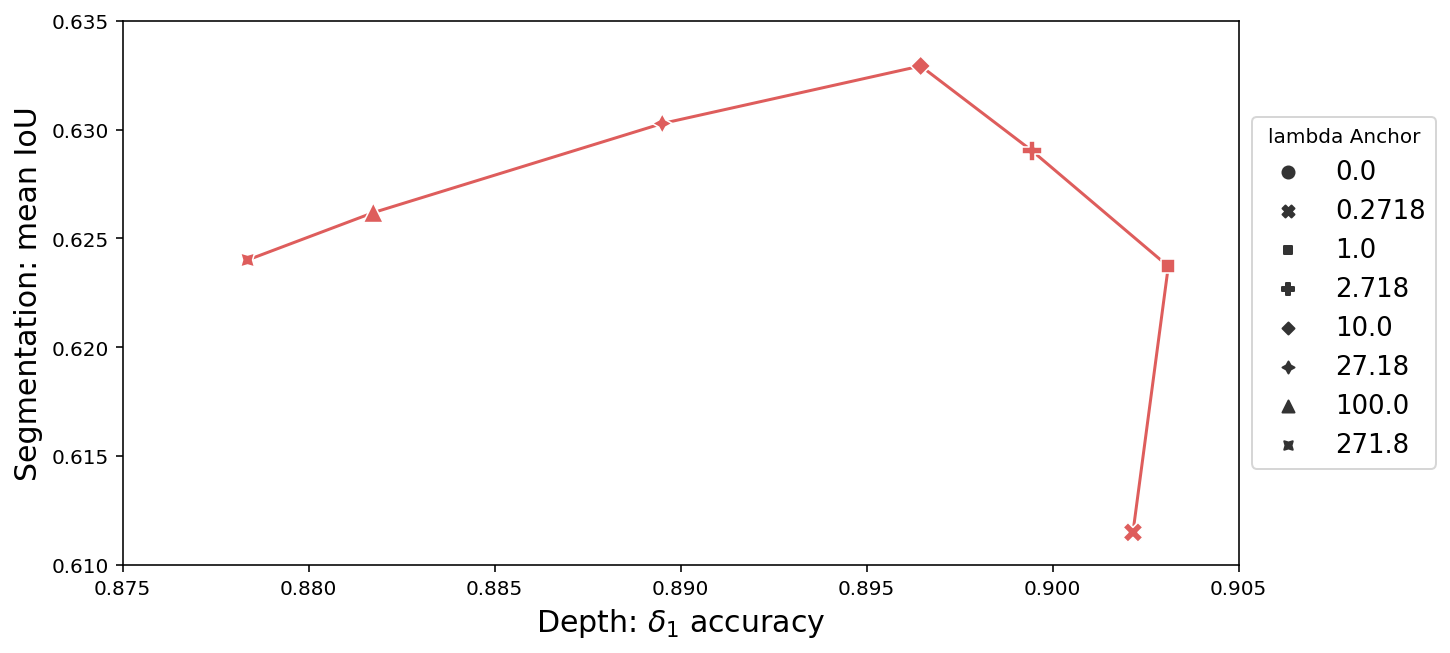} 
    \caption{Performance frontier (Segmentation vs. Depth)}
    \label{fig:frontier_plot-segmentation_vs_depth}
  \end{subfigure}

  \caption{
    \textbf{Analysis of the anchoring loss.}
    (a) Trade-off between cross-modal alignment and cross-scene discernibility, controlled by $\lambda_{anchor}$. The x-axis measures alignment (cosine sim of $<$RGB, Depth$>$) and the y-axis measures discernibility (1 - cosine similarity of distinct RGB scenes) on ScanNet. Frozen DINOv2 (light blue) is discriminative but poorly aligned.
    (b) To pick a value for $\lambda_{anchor}$, we examine its effect on linear-head prediction performance, from Omnivorous features of RGB images, on Depth (NYUv2) and Segmentation (Cityscapes). We omit the datapoint for $\lambda_{anchor} = 0$ located at $(x=0.732, y=0.356)$ for clarity, as it was too far below the remaining datapoints.\vspace{-2mm}
  }
  \label{fig:lambda-frontier}
\end{figure}

Figure \ref{fig:lambda-frontier} visualizes this frontier. The frozen DINOv2 baseline (light blue cross) exhibits high cross-scene discernibility (0.80) but suffers from poor cross-modal alignment (0.28), confirming our initial observations.

Our adapted features create a clear Pareto frontier. By varying $\lambda_{anchor}$, we can navigate this trade-off. Low values of $\lambda_{anchor}$ (e.g., 1.0) yield excellent cross-modal alignment (approaching 0.70 for dense features) but at the cost of reduced discriminative power, as the features drift significantly from the original encoder's semantic space.

Conversely, as $\lambda_{anchor}$ increases (e.g., to 10.0 or 100.0), the anchoring loss dominates. This pulls the adapted features back towards the frozen ones, recovering most of the original cross-scene discernibility but sacrificing the alignment gains. This result confirms that $\lambda_{anchor}$ acts as a ``knob'' to tune the desired balance.

\begin{table}[]
\centering
\caption{\textbf{Ablating modality mixup.} We vary $\alpha_{max}$ which controls the degree of blending between modalities during training. We report linear-probe performance on (i) classification (ImageNet accuracy using the TOK feature, without intermediate layers), (ii) depth prediction ($\delta_1$ on NYUv2), and (iii) segmentation (mean IoU on Cityscapes). We also report (iv) 3D correspondence (percentage of correct keypoints at threshold 0.0 on NAVI), which is assessed directly from features without a linear probe.}
\label{tab:mixup_ablation}
\begin{tabular}{lcccc}
\toprule
$\alpha_{max}$ & classif. $\uparrow$ & depth $\uparrow$   & segment. $\uparrow$ & 3D corresp. $\uparrow$ \\ \midrule
0              & 0.831  & \textbf{0.899} & 0.624  & 28.40  \\
0.25           & 0.834  & 0.898 & 0.630   & 28.96 \\
0.5            & 0.834  & 0.896 & \textbf{0.632}  & 29.00 \\
0.75           & 0.834  & 0.894 & \textbf{0.632}  & \textbf{29.04} \\
1.0            & \textbf{0.835}  & 0.891 & \textbf{0.632} & 29.03 \\ \bottomrule
\end{tabular}
\vspace{-2mm}
\end{table}

\noindent\textbf{Training data.} We ablate the mixup hyperparameter $\alpha_{max}$ which controls the degree to which the modalities are blended during training (see Table \ref{tab:mixup_ablation}). While depth prediction is an outlier, the performance on all other tasks continues to increase up to $\alpha_{max}=1$, which implements full-spectrum blending of the three modalities. Our default value $\alpha_{max}=0.5$ was chosen to balance across the tasks.

\noindent\textbf{Alternative parameterizations.} See Appendix \ref{app:extended_ablations} for results on: (i) using an alternative foundation model (TIPS \cite{tips_paper}) as the teacher network, (ii) learning an adapter on top of the teacher rather than adapting its final layers, and (iii) how many layers to freeze.%

\begin{figure}
    \centering
    \begin{subfigure}{0.9\linewidth}
        \centering
        \includegraphics[width=\linewidth]{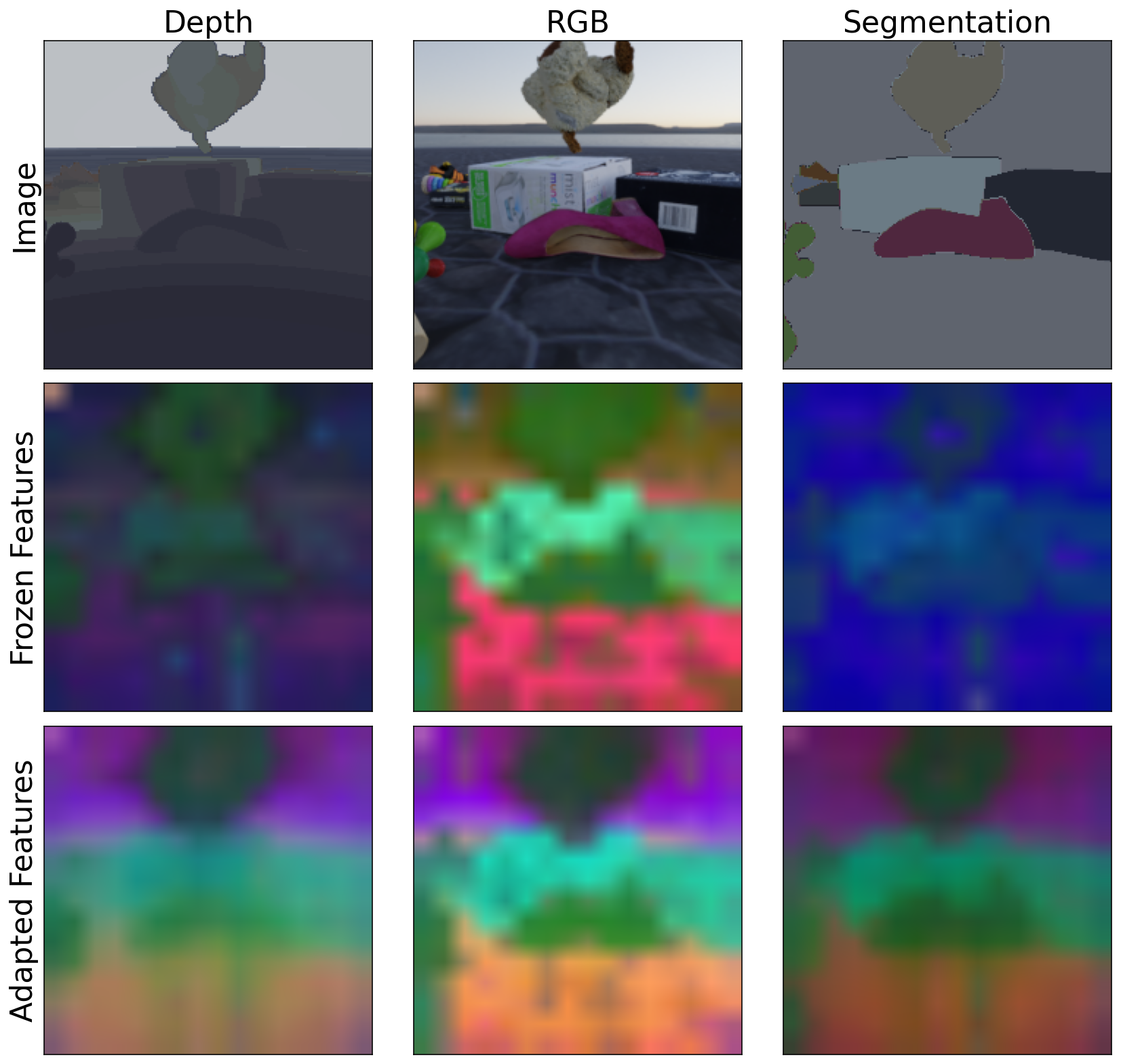}
        \caption{MOVi}
        \label{fig:wide_sub1}
    \end{subfigure}
    \hfill
    \begin{subfigure}{0.9\linewidth}
        \centering
        \includegraphics[width=\linewidth]{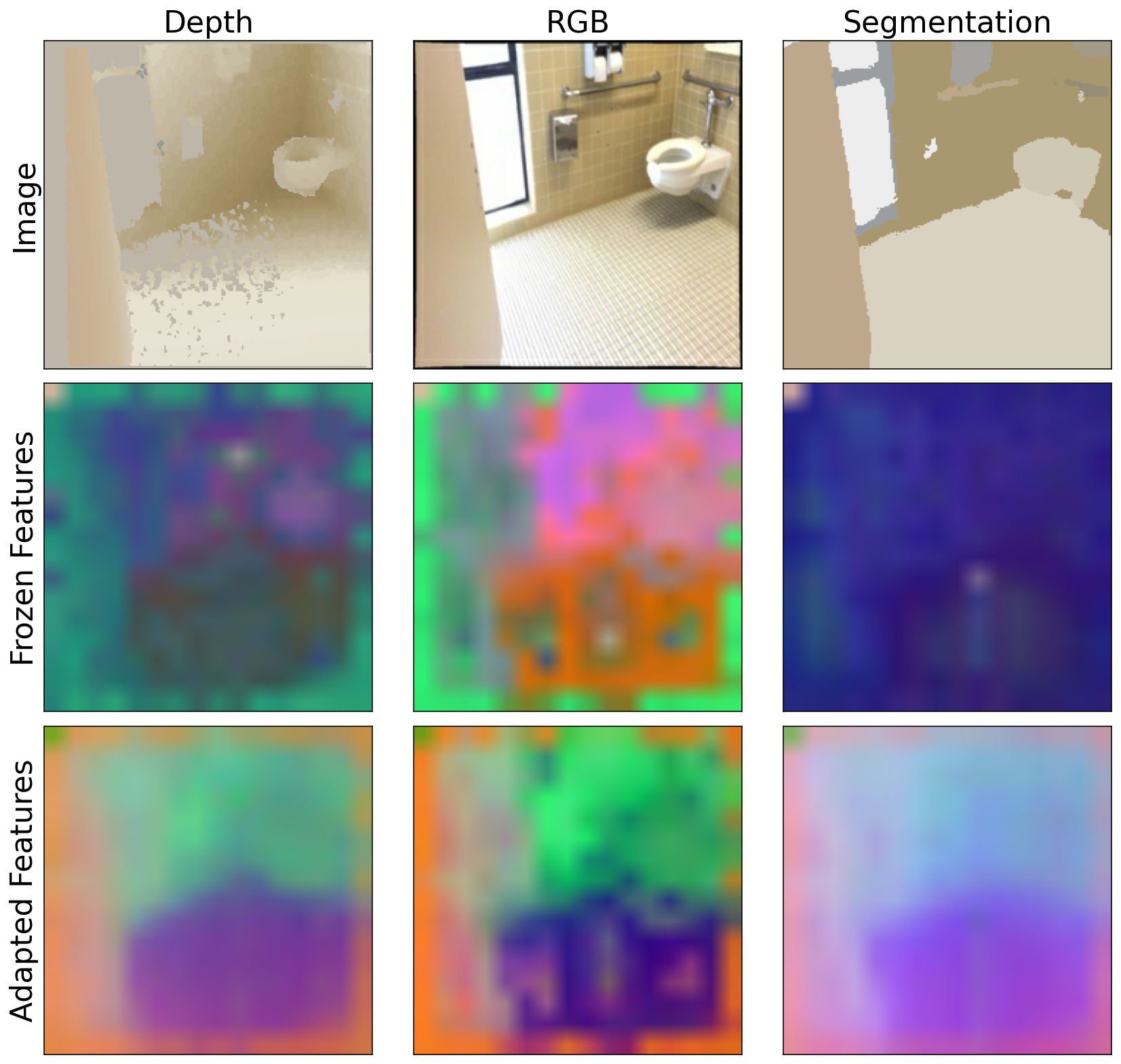}
        \caption{ScanNet}
        \label{fig:wide_sub2}
    \end{subfigure}
    \caption{PCA visualizations of frozen (DINO ViT-B/14) and adapted (Omnivorous ViT-B/14) features on two scenes.}
    \label{fig:pca_viz}
\end{figure}

\section{Discussion}

To verify the alignment of our learned representations, we visualize the top three Principal Components of the features. Figure~\ref{fig:pca_viz} presents two examples. The middle rows (``Frozen features'') illustrate the baseline DINOv2 output, where the feature maps for RGB, Depth, and Segmentation exhibit distinct color distributions, indicating that they occupy disjoint subspaces. In contrast, the bottom rows (``Adapted features'') demonstrate the effectiveness of our method: the feature maps for Depth and Segmentation align closely with the RGB features, sharing consistent colors and structural details. This qualitative evidence confirms that our adapter unifies the modalities into a shared semantic space without discarding spatial geometry.

\rk{Point to Appendix: train with a different vision model (TIPS). Train different sizes of DINOv2.}

\paragraph{Future work.} While in this work we focus on adapting a pre-existing feature space for cross-modality alignment, an interesting direction for exploration would be to align visual modalities while pre-training an encoder rather than post-hoc. This may unlock deeper benefits than fine-tuning the final layers of an existing model. In terms of potential downstream uses, having shown the benefits to cross-modal retrieval and depth prediction, we expect generative applications like monocular image-to-depth to benefit from conditioning on Omnivorous representations.

\paragraph{Limitations.} DINOv2 undergoes high-resolution fine-tuning as a final training step. It is unclear whether this step would be required after training Omnivorous DINO too.

\paragraph{Conclusion.} Distilling from DINOv2, we have shown that our Omnivorous approach can exceed DINOv2's performance on \textit{all} 3D-relevant tasks in the Probe3D framework, semantic tasks such as classification, and  cross-modal alignment. Our modality-agnostic encoder can also generalize to unseen visual modalities,  paving the way for a more foundational vision model.

\section*{Acknowledgments}
We thank Kevis-Kokitsi Maninis for help with the evaluations, and Goker Erdogan for comments on the draft.

{
    \small
    \bibliographystyle{ieeenat_fullname}
    \bibliography{main}
}

\clearpage
\setcounter{page}{1}
\maketitlesupplementary

This Appendix is organized into two broad sections: Section \ref{app:training_data} describes our training and evaluation framework, while Section \ref{app:extended_results} extends the results of the main paper.

\section{Training and Evaluation Details}
\label{app:training_data}

We first present a summary of our training configuration in Table \ref{tab:omni_dino_config}. Next, we describe our training data pipeline in Section \ref{app:data_pipeline}. Finally, we elaborate on our evaluation protocols in Section \ref{app:evaluation_protocols}.

\begin{table*}[h]
    \centering
    \caption{Training Configuration for Omnivorous DINO}
    \label{tab:omni_dino_config}
    \begin{tabular}{rl}
        \toprule
        \textbf{Category} & \textbf{Details} \\
        \midrule
        \multirow[t]{2}{*}{\textbf{Architecture}} & DINOv2 ViT-B/14 (173M parameters) \\
        & Layers 0--7 are frozen, 8--11 are fine-tuned. \\
        \textbf{Optimizer} & AdamW with learning rate $1 \times 10^{-4}$ \\
        \textbf{Compute} & TPU v4 ($4\times4\times4$) for 20,000 steps, with a total runtime of 1 hour 14 minutes \\
        \textbf{Batch Size} & 512 (Global) \\
        \textbf{Datasets} & ScanNet \cite{dai2017scannet}, TartanAir \cite{wang2020tartanair}, Hypersim \cite{roberts:2021}, MOVi \cite{greff2022kubric}, PointOdyssey \cite{zheng2023point}, DynamicReplica \cite{karaev2023dynamicstereo} \\
        \multirow[t]{5}{*}{\textbf{Preprocessing}} & $224 \times 224$ resolution (RGB--bilinear resize; Depth \& Seg--nearest neighbor; center crop to square) \\ 
        & Photometric Augmentation if training (RGB) \\ 
        & Colorization (Depth and Seg) \\ 
        & Normalization using ImageNet-1k mean and std (RGB, Depth, Seg) \\ 
        & Modality Mixup ($\alpha_{max}=0.5\text{ if training else }0.0$) \\
        
        \bottomrule
    \end{tabular}
\end{table*}

\subsection{Data Pipeline}
\label{app:data_pipeline}

Below we elaborate on all elements of the training-data processing steps previously introduced in Section \ref{sec:method_data}.

\subsubsection{Photometric Augmentation (RGB)}

The photometric augmentation pipeline applies a sequence of standard distortions to the RGB image to encourage robustness against lighting variations and color shifts. The pipeline first adjusts brightness by adding a delta sampled from $[-0.1,0.1]$. This is followed by a saturation adjustment, where the image is scaled by a factor drawn from $[0.8,1.2]$. Next, the hue is shifted by a delta within the range $[-0.03,0.03]$. Finally, the contrast is scaled by a factor sampled from $[0.8,1.2]$. All random scalars are sampled independently for each distortion type per image instance.

\subsubsection{Colorization (Depth \& Segmentation)}

Using standard colormaps (e.g., grayscale or jet) for the Depth and Segmentation images would allow the encoder to shortcut the alignment task by exploiting low-level channel statistics, thus learning modality-specific features. To counter this, we employ a \emph{natural colorization} strategy.

We define a transformation $\Phi(x_m^{\text{raw}}, x_r^{\text{aug}})$ that re-renders the scalar structural map $x_m^{\text{raw}}$ (e.g., depth) using the chromatic distribution of the corresponding RGB image $x_r^{\text{aug}}$. This process creates ``hard positives'' for the contrastive objective: by forcing the structural map to share the same color histogram as the RGB image, we deny the network the ability to distinguish or align modalities based on superficial color signals. Consequently, the encoder must attend to the shared geometric content to solve the alignment task.

\begin{algorithm}[t]
\caption{Natural Colorization}
\label{alg:natural_colorization}
\begin{algorithmic}[1]
\Require Scalar map $x^{\text{raw}}_m \in \mathbb{R}^{H \times W}$
\Statex \hskip2.5em where $m \in \{\text{Depth}, \text{Segmentation}\}$
\Require Augmented RGB image $x^{\text{aug}}_{r} \in \mathbb{R}^{H \times W \times 3}$
\Require Number of bins $B=64$, kernel size $K=5$,
\Statex \hskip2.5em constant $\epsilon=10^{-6}$
\Ensure Colorized map $x_m \in \mathbb{R}^{H \times W \times 3}$

\Statex
\State \textbf{Step 1: Normalization and Discretization}
\State $x^{\text{norm}}_m \leftarrow \frac{x^{\text{raw}}_m - \min(x^{\text{raw}}_m)}{\max(x^{\text{raw}}_m) - \min(x^{\text{raw}}_m) + \epsilon}$ \Comment{Normalize modality $m$ to $[0, 1]$}
\For{each pixel $(u, v)$}
    \State $b_{u,v} \leftarrow \text{clip}(\lfloor x^{\text{norm}}_m[u,v] \cdot B \rfloor, 0, B-1)$ \Comment{Compute bin indices}
\EndFor

\Statex
\State \textbf{Step 2: Palette Accumulation} \Comment{Aggregates $x^{\text{aug}}_{r}$ stats per bin}
\State Initialize $S \in \mathbb{R}^{B \times 3}$ and $N \in \mathbb{R}^{B}$ with zeros
\For{each pixel $(u, v)$}
    \State $k \leftarrow b_{u,v}$
    \State $S[k] \leftarrow S[k] + x^{\text{aug}}_{r}[u,v]$
    \State $N[k] \leftarrow N[k] + 1$
\EndFor

\Statex
\State \textbf{Step 3: Palette Smoothing} \Comment{Fills gaps via 1D convolution}
\State Define uniform kernel $w \in \mathbb{R}^K$ where $w_i = 1$
\State $\tilde{S} \leftarrow \text{Convolve1D}(S, w)$
\State $\tilde{N} \leftarrow \text{Convolve1D}(N, w)$

\Statex
\State \textbf{Step 4: Palette Normalization}
\For{$k \in \{0, \dots, B-1\}$}
    \State $\mathcal{P}[k] \leftarrow \tilde{S}[k] / (\tilde{N}[k] + \epsilon)$ \Comment{Compute avg color per bin}
\EndFor

\Statex
\State \textbf{Step 5: Image Re-rendering}
\For{each pixel $(u, v)$}
    \State $x_m[u,v] \leftarrow \mathcal{P}[b_{u,v}]$ \Comment{Map bins to palette colors}
\EndFor
\State \Return $x_m$
\end{algorithmic}
\end{algorithm}

See Algorithm \ref{alg:natural_colorization} for a pseudocode description of our natural colorization $\Phi$. Formally, we normalize $x_m^{\text{raw}}$ to $[0, 1]$ and discretize it into $B=64$ intensity bins (step 1). Let $b_{u,v} \in \{0, \dots, B-1\}$ denote the bin index of pixel $(u,v)$ in $x_m^{\text{raw}}$. We construct a scene-specific natural color palette $\mathcal{P} \in \mathbb{R}^{B \times 3}$ by aggregating the RGB colors corresponding to each structural intensity bin (step 2). The accumulated color sum $\mathbf{S}_k$ and pixel count $\mathbf{N}_k$ for bin $k$ are computed as:
$$\mathbf{S}_k = \sum_{u,v} \mathbf{1}[b_{u,v} = k] \cdot x_r^{\text{aug}}(u,v), \quad \mathbf{N}_k = \sum_{u,v} \mathbf{1}[b_{u,v} = k]$$

To ensure continuity, we apply a 1D smoothing convolution to $\mathbf{S}$ and $\mathbf{N}$ using a kernel of size 5 (step 3). The final palette value for bin $k$ is $\mathcal{P}_k = \tilde{\mathbf{S}}_k / (\tilde{\mathbf{N}}_k + \epsilon)$, where $\epsilon$ is set to $1e-6$ for numerical stability. The colorized map $x_m$ is generated by mapping each pixel in the raw map to its corresponding palette entry: $x_m(u,v) = \mathcal{P}_{b_{u,v}}$.

\subsubsection{Normalization (RGB, Depth, \& Segmentation)}
We use the ImageNet-1k mean pixel value $(0.485, 0.456, 0.406)$ and standard deviation $(0.229, 0.224, 0.225)$ to standardize all [0,1] images. 

\subsubsection{Modality Mixup}

While natural colorization forces the encoder to focus on structure, it leaves the depth and segmentation maps stripped of textured. Due to this domain gap, the model may struggle to relate geometric shapes to rich photometric cues. To bridge the gap, we use modality mixup. By stochastically blending the colorized structural maps with the original RGB image, we span a continuous ``modality spectrum'' that interpolates between pure geometry (Depth/Segmentation) and pure texture (RGB). This exposes the encoder to a smooth space of inputs, encouraging it to learn representations that are invariant to the ratio of texture-to-structure, rather than overfitting to discrete modality tokens.

Let $x_m$ be the naturally colorized map for modality $m \in \{\text{Depth}, \text{Segmentation}\}$ (from Algorithm \ref{alg:natural_colorization}) and $x^{\text{aug}}_{r}$ be the photometrically augmented RGB image. We generate the final mixed input $x^{\text{mixup}}_m$ via convex combination:$$x^{\text{mixup}}_m = (1 - \alpha_m) x_m + \alpha_m x^{\text{aug}}_{r}$$where the mixing coefficient $\alpha_m$ is sampled uniformly from the range $[0, \alpha_{\text{max}}]$ independently for each training example. We set $\alpha_{\text{max}} = 0.5$ to ensure the structural signal remains dominant while re-introducing sufficient texture to facilitate alignment. This strategy effectively constructs a "continuous bridge" between modalities, preventing the feature space from fragmenting into disjoint islands of geometry and texture.

\subsection{Evaluation Protocols}
\label{app:evaluation_protocols}

We adopt the protocols established by DINOv2 \cite{oquab2024dinov} or Probe3D \cite{El_Banani_2024_CVPR} wherever possible. We elaborate all details in the following subsections for completeness. 

\subsubsection{Cross-Modal Retrieval} For all datasets (ScanNet, MOVi, TartanAir), inputs are resized to 224×224 (using bilinear interpolation for RGB and nearest-neighbor for depth/segmentation) followed by a center crop. Single-channel structural inputs (depth and segmentation) are tiled to 3 channels and normalized using standard ImageNet statistics ($\mu=[0.485,0.456,0.406], \sigma=[0.229,0.224,0.225]$) after scaling pixel values to $[0,1]$. Features are extracted using the frozen DINOv2 backbone and our adapter, applying $L_2$ normalization to the final embeddings.

We compute pairwise cosine similarity between the query and gallery sets. To handle large-scale evaluation efficiently, similarity matrices are computed in batches of 2048. The rank for a given query is determined by counting the number of gallery items with a similarity score strictly greater than or equal to the ground-truth pair's score (using a numerical stability threshold $\epsilon=10 
^{-6}$). As our evaluation setup assumes a strict one-to-one mapping between modalities (i.e., exactly one positive match per query), the Mean Average Precision (mAP) reported is equivalent to the Mean Reciprocal Rank (MRR). We average results over all six directed modality pairs.

\subsubsection{Monocular Depth Estimation}
\label{app:evaluation_protocols_depth}

\textbf{Data and Preprocessing.} We evaluate on NYUv2 \cite{Silberman:ECCV12} and NAVI Probe3D \cite{El_Banani_2024_CVPR}. Unlike the classification or retrieval tasks which often resize inputs to a standard 224×224, we perform evaluation on high-resolution images to preserve geometric details (i.e., 480×640 for NYUv2, 512×512 for NAVI). To process these variable resolutions with a ViT backbone trained on fixed patch sizes, we employ a ``pad-to-patch" strategy: images are first center-cropped to the target resolution and then padded to the nearest multiple of the patch size (p=14). This allows the frozen backbone to process the dense grid of patches without interpolation artifacts. Standard photometric distortions and random rotations are applied during training, while horizontal flipping is used for test-time augmentation.

\textbf{Decoder Architectures.} We investigate the expressivity of our learned features using two distinct decoder heads.

\begin{itemize}
    \item \textbf{Linear Head:} A lightweight baseline that projects the final layer's patch tokens directly to depth bins using a single linear layer. The output is bilinearly upsampled to the input resolution. This setup tests the explicit geometric information present in the final semantic embedding.
    
    \item \textbf{DPT Head:} A Dense Prediction Transformer (DPT) decoder that aggregates intermediate features from the backbone. Specifically, we gather tokens from layers {3, 6, 9, 12} (for the ViT-B/14 variant), fuse them using valid convolutions and upsampling blocks to recover high-resolution details. This head evaluates the backbone's ability to provide multi-scale hierarchical features suitable for dense prediction.
\end{itemize}

\textbf{Training Objective.} Both heads are trained (while keeping the backbone frozen) to classify pixels into 256 depth bins. We minimize a combined objective consisting of a Scale-Invariant Gradient Loss (sigloss)  to enforce global structural consistency and an edge-aware gradient loss to sharpen local discontinuities. We train for 50,000 steps using AdamW with a compound learning rate schedule (constant, piecewise constant, and linear warmup).

\subsubsection{Semantic Segmentation}

\textbf{Data and Preprocessing.} We evaluate semantic segmentation on ADE20k, Cityscapes, and Pascal VOC. During training, we employ standard data augmentation techniques: input images undergo random resizing (ratio range $[0.5, 2.0]$), random horizontal flipping, and photometric distortion.  The images are then randomly cropped to a fixed resolution of $512\times512$.

\textbf{Evaluation Protocol.} Unlike the monocular depth evaluation which processes full images via padding, our segmentation evaluation employs a sliding window protocol to handle high-resolution inputs (e.g., Cityscapes) without downsampling artifacts. We perform inference on $512\times512$ crops with a stride of 341 pixels.  Predictions from overlapping windows are averaged (mean logits) before the final argmax.

\textbf{Decoder Architectures.} We utilize the same two decoder configurations---Linear and DPT---as described in the Monocular Depth Estimation section (Appendix \ref{app:evaluation_protocols_depth}). The backbone remains frozen as before. The only modification is the final projection layer, which maps to $K$ semantic classes (e.g., $K=150$ for ADE20k) instead of depth bins.

\textbf{Optimization.} We train for 40,000 steps with a batch size of 16. We use the AdamW optimizer with a weight decay of $10^{-4}$. The learning rate follows a polynomial decay schedule (power $1.0$) combined with a linear warmup for the first 1,500 steps. Performance is measured using the Mean Intersection-over-Union (mIoU), computed by aggregating confusion matrices over the entire validation set.

\subsubsection{Multiview Correspondence}

\textbf{Data and Preprocessing.} We evaluate 3D feature correspondence using the NAVI dataset. Image pairs are resized to $224\times224$. We extract feature maps from the encoder, which correspond to a $16\times16$ grid of patches (given the patch size $p=14$). We do not employ a trained prediction head for this task; instead, we evaluate the raw feature representations directly.

\textbf{Matching Protocol.} For a given image pair, we compute the pairwise cosine similarity matrix between the flattened spatial tokens ($N=256$) of the source and target views. We determine the predicted correspondence for each token by selecting the nearest neighbor (argmax of cosine similarity) in the other view. We evaluate bidirectional matches.

\textbf{Metric: PCK@0.} We report the Percentage of Correct Keypoints (PCK) at a strict threshold of $0.0$. Since our evaluation operates on the discrete $16\times16$ token grid, a threshold of $0.0$ requires the predicted token index to exactly match the ground-truth token index (i.e., the predicted patch must be the exact same patch as the ground truth). We generally report performance using the final layer's features. That said, we also include an ablation (Table \ref{tab:multiview_correspondence}) measuring 3D correspondence across all fine-tuned Omnivorous ViT blocks (i.e., the last four).

\subsubsection{Linear Probe Classification}

\textbf{Architecture and Training.} To assess the linear separability of the learned representations, we train a linear classifier on top of the frozen backbone. We attach a single linear layer (projecting from the feature dimension $D$ to the number of classes $K=1000$) to the extracted features. The linear layer is trained to minimize the weighted softmax cross-entropy loss, while the backbone remains frozen.

\textbf{Evaluation Protocol.} We evaluate on ImageNet-1k \cite{deng2009imagenet}, reporting top-1 accuracy on the validation split. We employ standard data augmentation during training (random resized crops and horizontal flips), while validation images are resized to 256 pixels and center-cropped to $224\times224$. We  and train the prober for 10 epochs, sweeping over a range of learning rates (base values: [0.15, 0.2, 0.5, 1.0, 2.0]) along with the nesterov optimizer. We report the best accuracy achieved across the base learning rates. We report results using both the CLS token embedding and the concatenation of the CLS token and the global average pooled (GAP) features.

\subsubsection{k-NN Classification}

\textbf{ImageNet.} We follow the standard DINO evaluation protocol for ImageNet-1k. We extract features for the training set (index) and validation set (query) using the frozen backbone. The images are preprocessed by resizing the shorter side to 256 pixels, taking a central $224\times224$ crop, and normalizing with ImageNet statistics. We employ weighted soft voting: for each query, we compute the cosine similarity with its $k$ nearest neighbors in the training set. These similarities are converted to weights using a softmax with temperature $\tau=0.07$. The class probabilities are summed across the neighbors, and the class with the highest aggregate probability is selected. We report the top-1 accuracy corresponding to the best $k$ swept over $\{5, 10, 20, 50, 100\}$.

\textbf{Transfer Datasets.} For iNaturalist, SOP, Google Landmarks v2 (GLDv2), RP2K, and Food2k, we perform "hard" k-NN classification (which is equivalent to Recall@1). We use the same image preprocessing as ImageNet ($256\rightarrow224$ center crop).
\begin{itemize}
    \item For \textbf{GLDv2}, we match queries from the test set against the distinct index set provided by the dataset ($N\approx761k$).
    \item For \textbf{iNaturalist, SOP, RP2K, and Food2k}, we follow the standard metric learning protocol where the test set serves as both the query and the index. We compute the nearest neighbor for each query from the index, excluding the query itself (self-match), and check if the retrieved class label matches the query label.
\end{itemize}

\subsubsection{Zero-Shot Modality Transfer}

\textbf{Protocol.} To assess the universality of the learned feature space, we design a strict transfer protocol. We train a depth estimation head (either Linear or DPT as before) using \textit{only} RGB images from the NYUv2 dataset. Once trained, we freeze the entire model (backbone + depth head) and evaluate it on the PACE dataset. This setup introduces a small domain shift and, crucially, a modality shift.

\textbf{Modalities.} We evaluate performance on three distinct input types:
\begin{itemize}
    \item \textbf{RGB:} Serves as the baseline. The model encounters a domain shift (NYUv2 $\rightarrow$ PACE) but the modality remains consistent with training.
    \item \textbf{Segmentation:} A modality seen by the Omnivorous backbone during pre-training, but \textit{never} seen by the depth head. To render these inputs compatible with the frozen backbone, segmentation maps are preprocessed using our Natural Colorization scheme (Algorithm 1) to match the spectral statistics of RGB images. Unlike the backbone pre-training stage, we do \textit{not} apply modality mixup during this evaluation.
    \item \textbf{NOCS (Normalized Object Coordinate Space):} NOCS maps represent dense coordinate fields rather than photometric data. It is a modality that is completely out-of-distribution; neither the Omnivorous backbone nor the depth head observes NOCS maps during training. The 3-channel coordinate maps are normalized using standard ImageNet RGB statistics before being fed into the model.
\end{itemize}

Success on NOCS and Segmentation inputs indicates that the encoder maps these diverse signals to a shared feature space that is interpretable by the RGB-trained head.

\section{Extended Results}
\label{app:extended_results}

\subsection{Diagnostic Metrics}

Expanding on Fig \ref{fig:motivation}, we report detailed cross-modal alignment and cross-scene discernibility metrics before and after Omnivorous training. Table \ref{tab:more_diagnostics} shows that our default checkpoint of Omnivorous DINO greatly improves cross-modal alignment while sacrificing some cross-scene discernibility (e.g., from 0.198 to 0.259 $<R_1, R_2>$ similarity on ScanNet). This echoes Fig \ref{fig:frontier-plot} which showed the trade-off as a function of our $\lambda_{anchor}$ loss weight.

\begin{table*}
    \centering
    \caption{\textbf{Diagnostic metrics:} we expand Fig 1, showing cross-modal alignment and cross-scene discernibility metrics across three datasets for both pretrained DINOv2 and the adapted Omnivorous model (at our default $\lambda_{anchor}=10)$). We denote the three modalities R, D, and S (RGB, Depth, and Segmentation, respectively). The metrics are computed without modality-mixup (i.e., $\alpha_{max} = 0$). For $<R_1, R_2>$, lower similarity is considered better.}
    \label{tab:more_diagnostics}
\begin{tabular}{lcccc|cccc}
\toprule
 & \multicolumn{4}{c}{DINOv2 ViT-B/14} & \multicolumn{4}{c}{Omnivorous ViT-B/14} \\
 dataset & $<R, D>$ & $<R, S>$ & $<D, S>$ & $<R_1, R_2>$ & $<R, D>$ & $<R, S>$ & $<D, S>$ & $<R_1, R_2>$ \\
\midrule
movi & 0.263 & 0.284 & 0.481 & 0.237 & 0.567 & 0.579 & 0.721 & 0.279 \\
scannet & 0.285 & 0.216 & 0.413 & 0.198 & 0.600 & 0.550 & 0.663 & 0.259 \\
tartanair & 0.345 & 0.359 & 0.543 & 0.172 & 0.607 & 0.603 & 0.736 & 0.223 \\
\bottomrule
\end{tabular}
\end{table*}

\subsection{3D Tasks}

We revisit all tasks from the Probe3D framework for the Omnivorous DINO ViT-B/14 checkpoint introduced in Sec \ref{sec:experiments}. We present two evaluations that were omitted in the main paper (normals estimation and multiview 3D correspondence), and add qualitative results for those already presented in the main paper (e.g., depth estimation and segmentation):

\subsubsection{Normals Estimation}
See Table \ref{tab:normals}. Omnivorous is consistently at par with DINOv2 across all metrics. 

\begin{table*}
    \centering
    \caption{\textbf{Downstream eval:} normals estimation using a DPT head.}
    \label{tab:normals}
\begin{tabular}{llcccccc}
\toprule
 &  & absrel $\downarrow$ & diff 11.25 $\uparrow$ & diff 22.50 $\uparrow$ & diff 30.00 $\uparrow$ & mean diff angle $\downarrow$ & rmse angle $\downarrow$ \\
dataset & model &  &  &  &  &  &  \\
\midrule
\multirow[t]{2}{*}{navi} & DINOv2 ViT-B/14 & 197.9 & 43.5 & 72.2 & 82.1 & 18.6 & 24.6 \\
 & Omnivorous ViT-B/14 & \textbf{197.8} & \textbf{43.6} & \textbf{72.3} & \textbf{82.2} & 18.6 & 24.6 \\
\cline{1-8}
\multirow[t]{2}{*}{nyuv2} & DINOv2 ViT-B/14 & 134.9 & 63.4 & 80.8 & 86.5 & 14.1 & 21.7 \\
 & Omnivorous ViT-B/14 & \textbf{134.1} & \textbf{63.5} & 80.8 & 86.5 & 14.1 & \textbf{21.6} \\
\cline{1-8}
\bottomrule
\end{tabular}
\end{table*}

\subsubsection{Multiview Correspondence}

See Table \ref{tab:multiview_correspondence}. While our model is consistently more 3D-consistent than the original DINOv2, the performance gap is a bit inconsistent with respect to the block where the features are taken from, i.e., there is no clear pattern of increasing/decreasing 3D-correspondence as a function of network depth. This merits future investigation.

\begin{table}
    \centering
    \caption{\textbf{Multiview correspondence:} we report the Percentage of Correct Keypoints ($\uparrow$) at the 0.0 level (i.e., only exact matches are counted). We measure correspondence for all the four blocks that are fine-tuned in the Omnivorous case, comparing them with their frozen DINOv2 counterparts.}
    \label{tab:multiview_correspondence}
    \begin{tabular}{lrrrr}
    \toprule
    block number & 9 & 10 & 11 & 12 \\
    model \\
    \midrule
    DINOv2 ViT-B/14 & 29.76 & 28.49 & 27.68 & 28.57 \\
    Omnivorous ViT-B/14 & 29.76 & \textbf{28.93} & \textbf{28.63} & \textbf{29.00} \\
    \bottomrule
    \end{tabular}
        
\end{table}

\subsubsection{Semantic Segmentation}
See Fig \ref{fig:seg_vis} \& \ref{fig:seg_vis_2} for a qualitative comparison between Omnivorous and DINOv2. We find that our model helps reduce over-segmentation, and is consistently more resilient to textural details in the input images.

\subsubsection{Monocular Depth}

\begin{figure*}
    \centering
    \begin{subfigure}{0.85\linewidth}
        \centering
        \includegraphics[width=\linewidth]{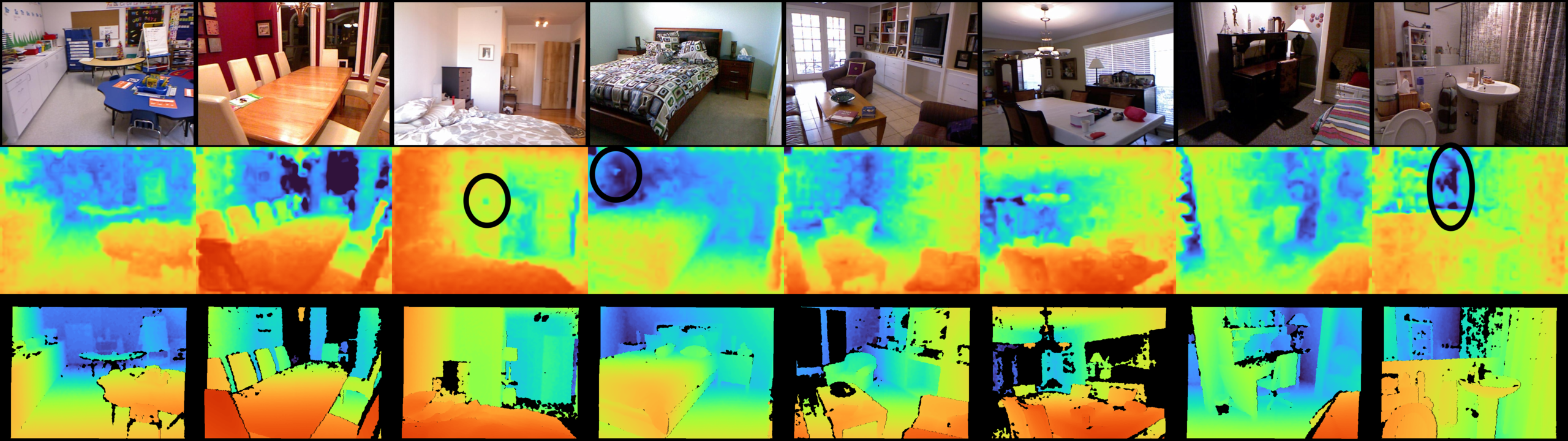}
        \caption{DINO ViT-B/14 depth prediction on NYUv2. Top: input images, middle: predictions, bottom: ground-truth.}
        \label{fig:wide_sub1}
    \end{subfigure}
    \hfill
    \begin{subfigure}{0.85\linewidth}
        \centering
        \includegraphics[width=\linewidth]{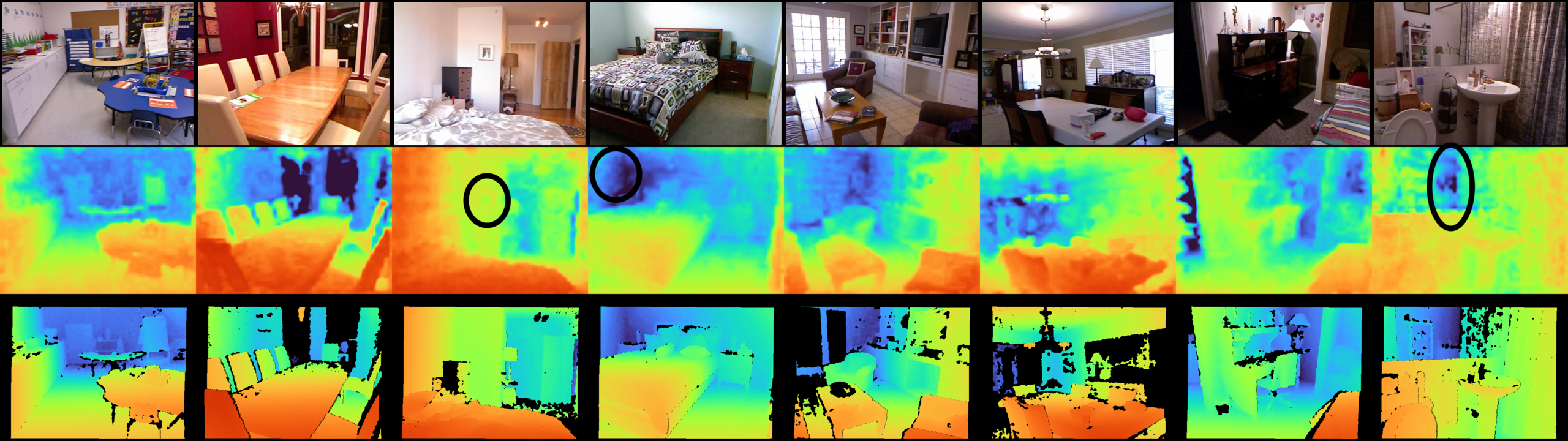}
        \caption{Omnivorous DINO ViT-B/14 depth prediction on NYUv2. Top: input images, middle: predictions, bottom: ground-truth.}
        \label{fig:wide_sub2}
    \end{subfigure}
        \hfill
    \begin{subfigure}{0.85\linewidth}
        \centering
        \includegraphics[width=\linewidth]{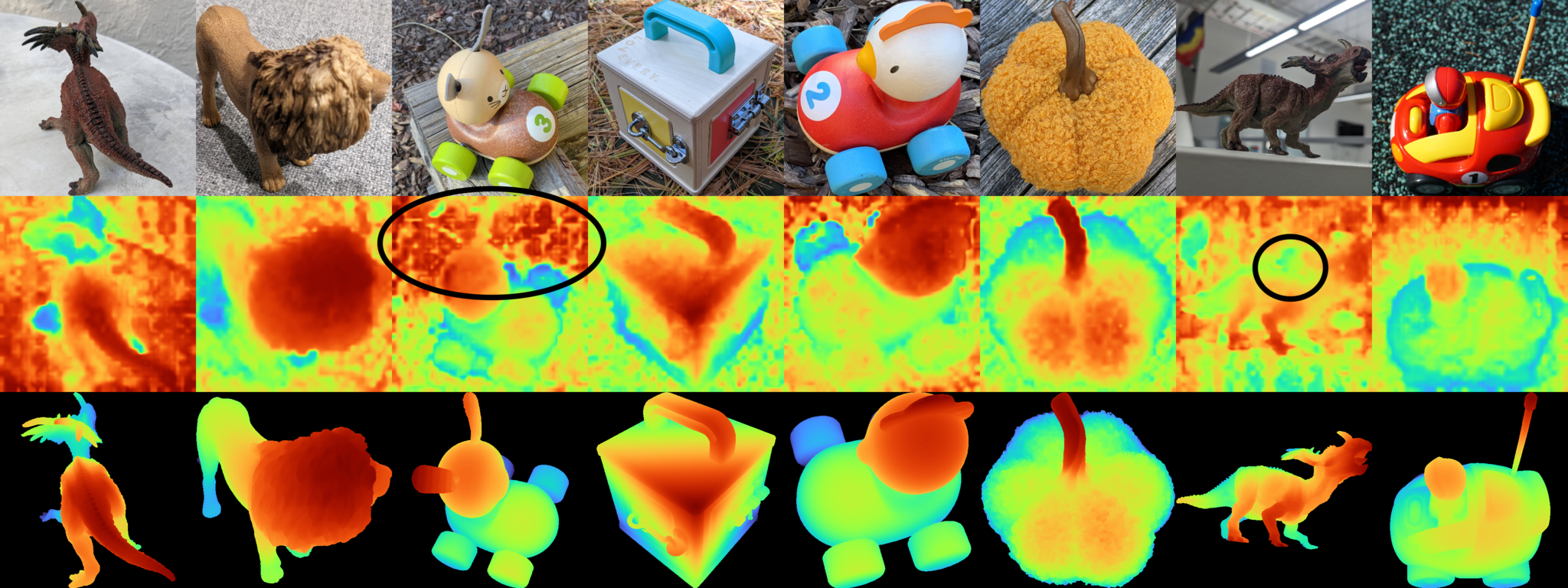}
        \caption{DINO ViT-B/14 depth prediction on NAVI Probe3D. Top: input images, middle: predictions, bottom: ground-truth.}
        \label{fig:wide_sub2}
    \end{subfigure}
    \hfill
    \begin{subfigure}{0.85\linewidth}
        \centering
        \includegraphics[width=\linewidth]{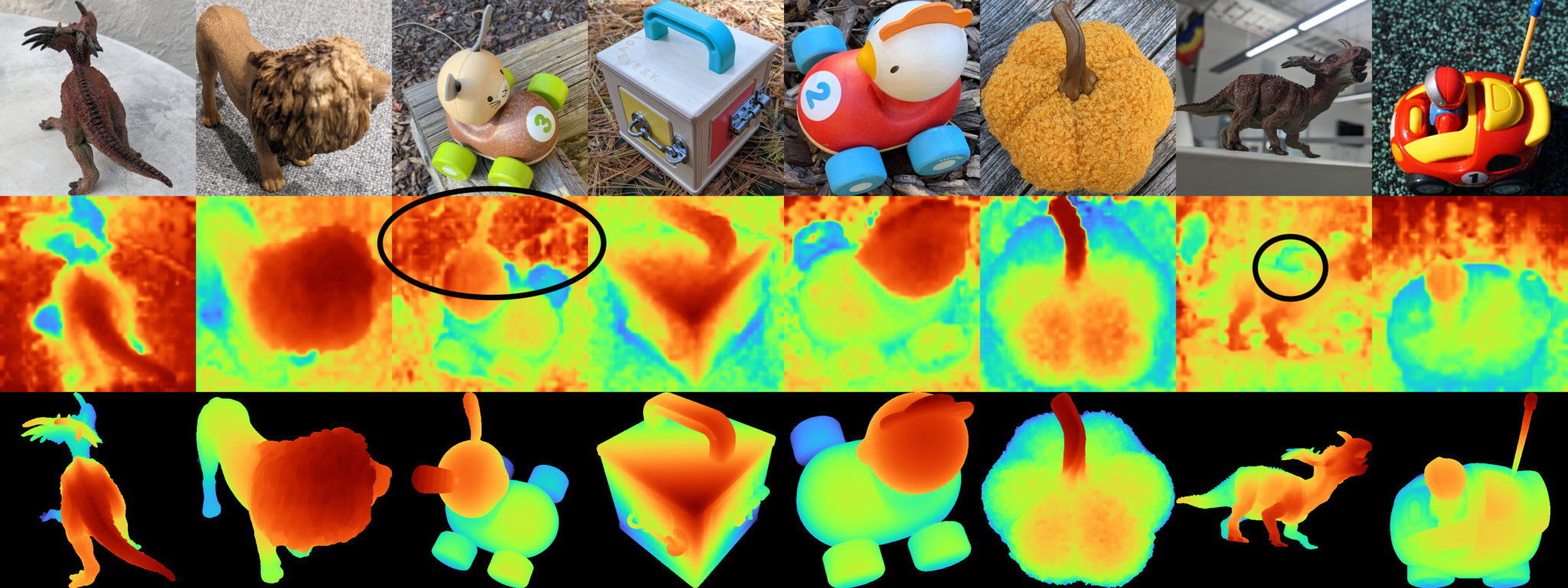}
        \caption{Omnivorous DINO ViT-B/14 depth prediction on NAVI Probe3D. Top: input images, middle: predictions, bottom: ground-truth.}
        \label{fig:wide_sub2}
    \end{subfigure}

    \caption{\textbf{Qualitative comparison (Omnivorous vs DINOv2) on depth prediction} using a linear head. Please compare a versus b, and c versus d. We highlight notable differences using a black oval.}
    \label{fig:depth_vis}
\end{figure*}

\begin{figure*}
    \centering
    \begin{subfigure}{0.7\linewidth}
        \centering
        \includegraphics[width=\linewidth]{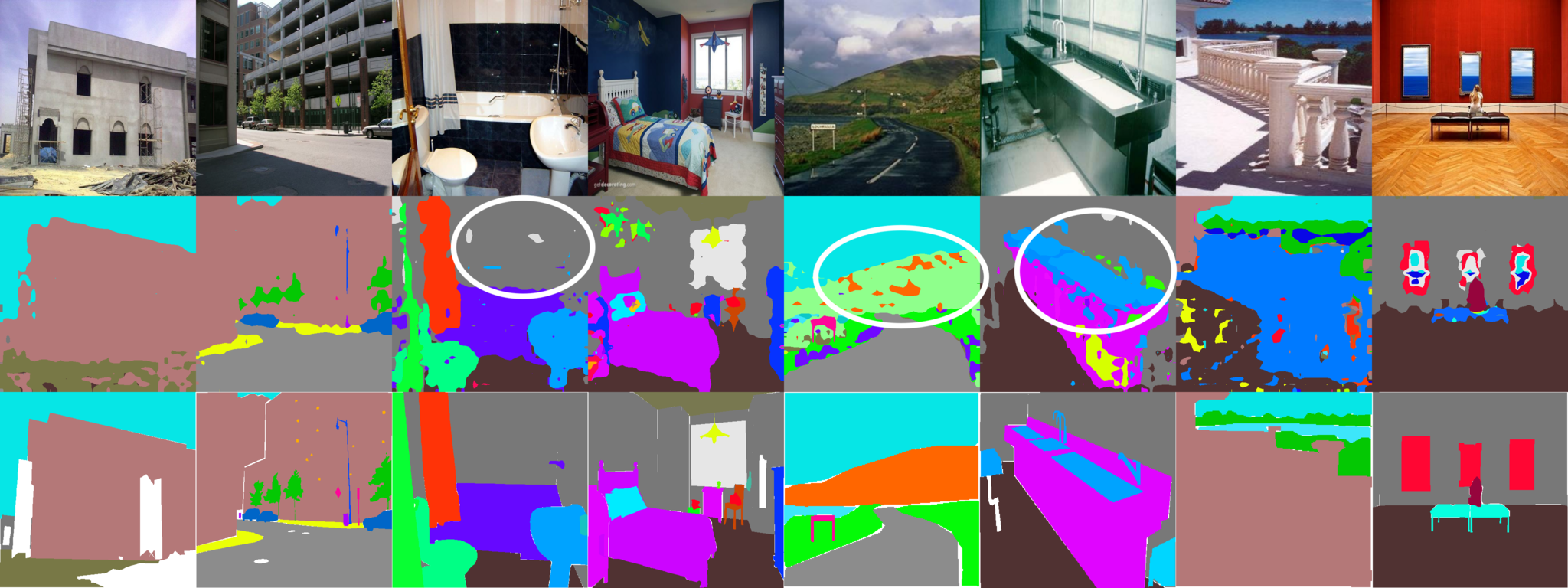}
        \caption{DINO ViT-B/14 segmentation prediction on ADE20k. Top: input images, middle: predictions, bottom: ground-truth.}
        \label{fig:wide_sub1}
    \end{subfigure}
    \hfill
    \begin{subfigure}{0.7\linewidth}
        \centering
        \includegraphics[width=\linewidth]{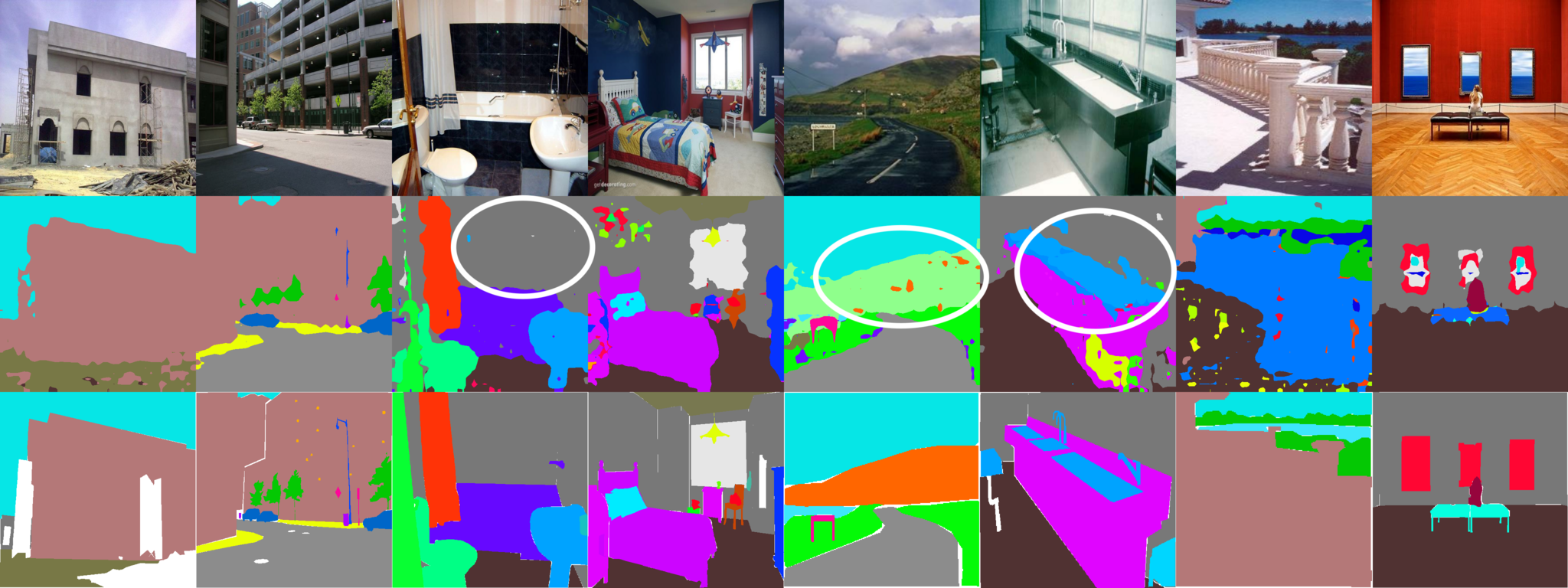}
        \caption{Omnivorous DINO ViT-B/14 segmentation prediction on ADE20k. Top: input images, middle: predictions, bottom: ground-truth.}
        \label{fig:wide_sub2}
    \end{subfigure}
        \hfill
    \begin{subfigure}{0.7\linewidth}
        \centering
        \includegraphics[width=\linewidth]{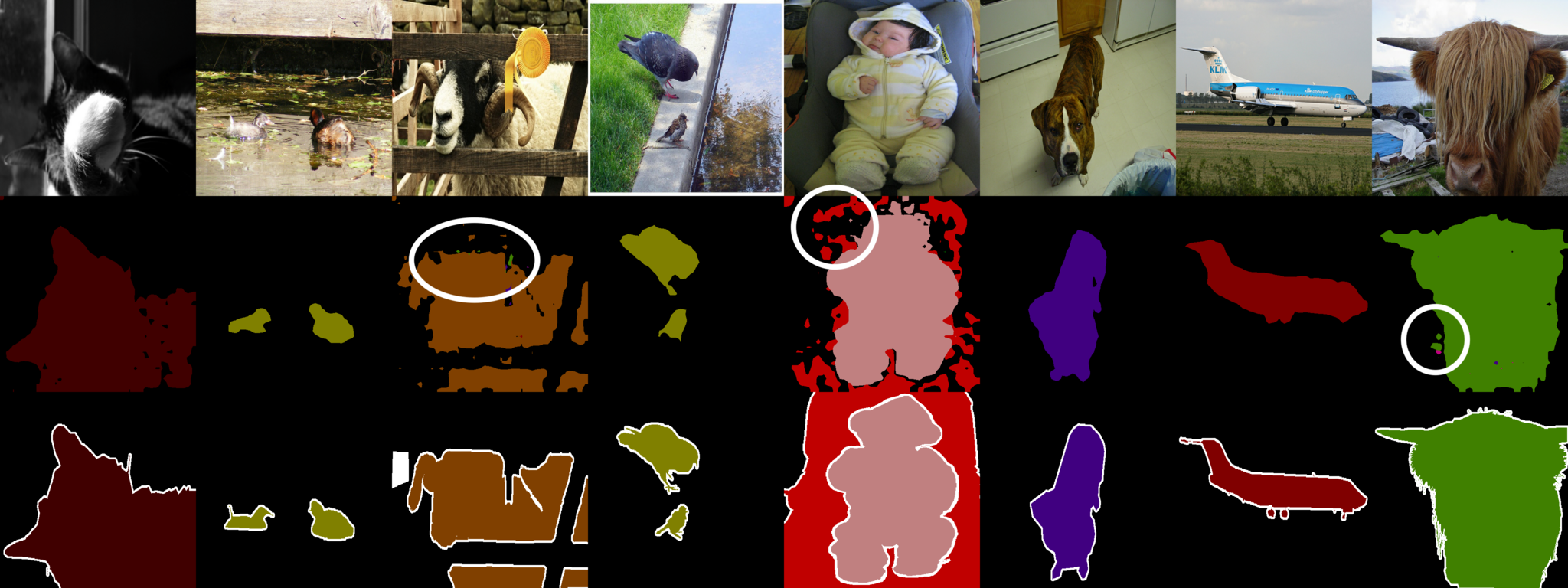}
        \caption{DINO ViT-B/14 segmentation prediction on Pascal VOC. Top: input images, middle: predictions, bottom: ground-truth.}
        \label{fig:wide_sub2}
    \end{subfigure}
    \hfill
    \begin{subfigure}{0.7\linewidth}
        \centering
        \includegraphics[width=\linewidth]{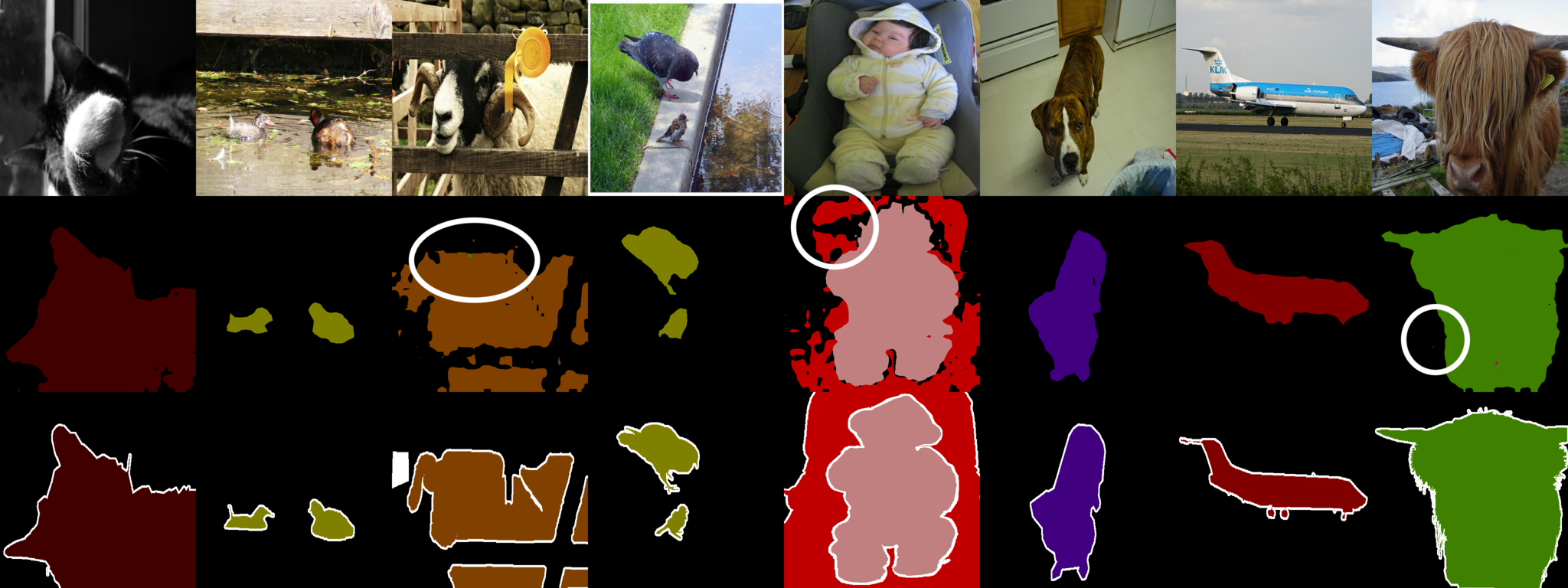}
        \caption{Omnivorous DINO ViT-B/14 segmentation prediction on Pascal VOC. Top: input images, middle: predictions, bottom: ground-truth.}
        \label{fig:wide_sub2}
    \end{subfigure}

    \caption{\textbf{Qualitative comparison (Omnivorous vs DINOv2) on segmentation prediction} using a linear head.  We highlight notable differences using a white oval.}
    \label{fig:seg_vis}
\end{figure*}

\begin{figure*}
    \centering
    \begin{subfigure}{0.7\linewidth}
        \centering
        \includegraphics[width=\linewidth]{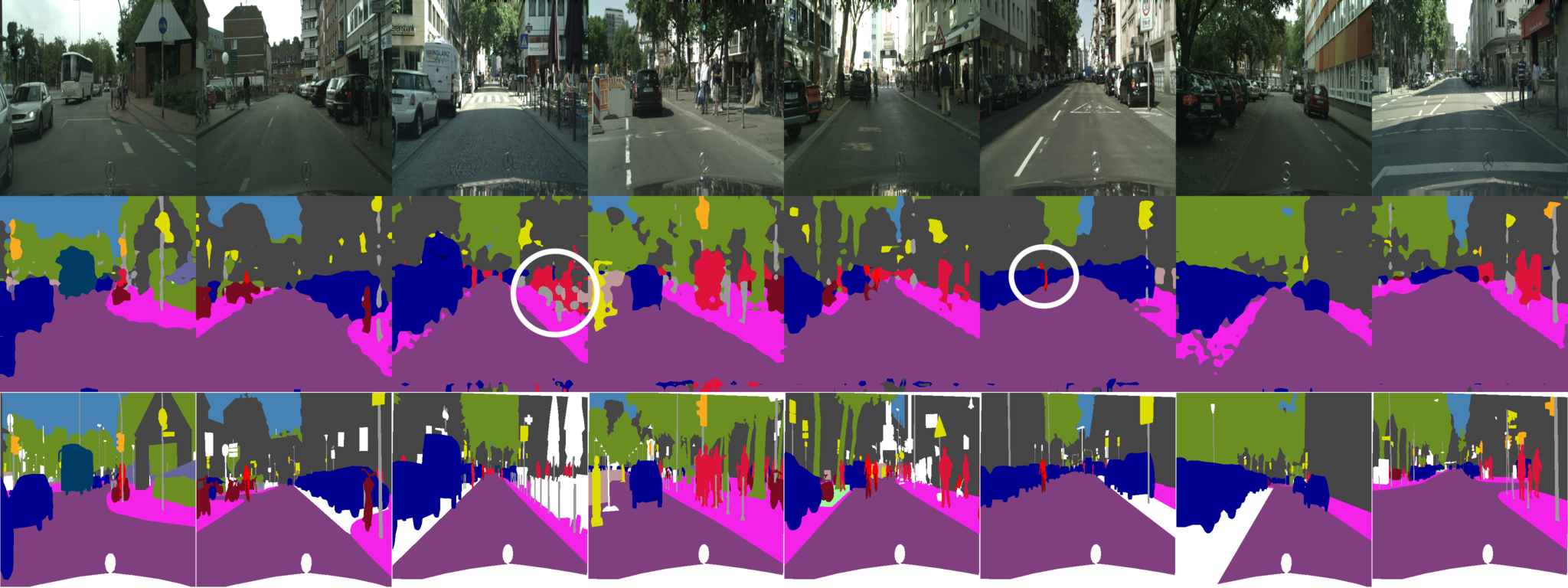}
        \caption{DINO ViT-B/14 segmentation prediction on Cityscapes. Top: input images, middle: predictions, bottom: ground-truth.}
        \label{fig:wide_sub1}
    \end{subfigure}
    \hfill
    \begin{subfigure}{0.7\linewidth}
        \centering
        \includegraphics[width=\linewidth]{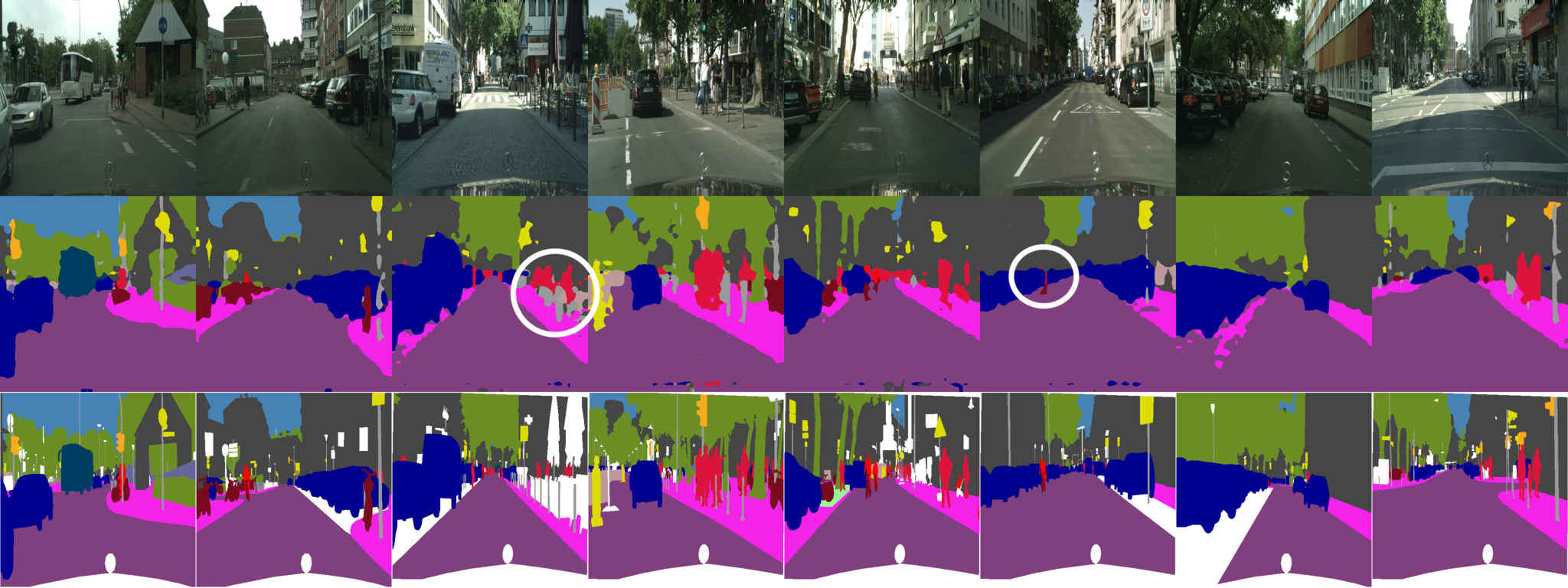}
        \caption{Omnivorous DINO ViT-B/14 segmentation prediction on Cityscapes. Top: input images, middle: predictions, bottom: ground-truth.}
        \label{fig:wide_sub2}
    \end{subfigure}
    \caption{\textbf{Qualitative comparison (Omnivorous vs DINOv2) on segmentation prediction (contd.)} using a linear head. We highlight notable differences using a white oval.}
    \label{fig:seg_vis_2}
\end{figure*}

See Fig \ref{fig:depth_vis} for a qualitative comparison between Omnivorous and DINOv2. As with predicted segmentations, we find that our model helps reduce high-frequency noise in the linear head's depth predictions. Our model performs consistently better on flat surfaces, and cases where a flat object is placed on a flat surface (e.g., a painting on the wall).

\subsection{Ablations}
\label{app:extended_ablations}

\subsubsection{TIPS instead of DINOv2}
\label{app:ablation_tips}

As TIPS \cite{tips_paper} shares the same ViT architecture as DINOv2, we can ``ablate'' our pretrained teacher by running Omnivorous training on TIPS instead of DINOv2. Two important distinctions are the shape of the position encoding parameter (TIPS uses $16 \times 16$ vs DINOv2's $37\times37$) and the number of CLS tokens (TIPS uses two while DINOv2 uses one). We train Omnivorous TIPS ViT-B/14 using the default $\alpha_{max}=0.5$, and freezing the first 8 blocks as we did for Omnivorous DINOv2.

\begin{figure}[htbp]
  \centering
  
  \begin{subfigure}[b]{\linewidth}
    \centering
    \includegraphics[width=\linewidth]{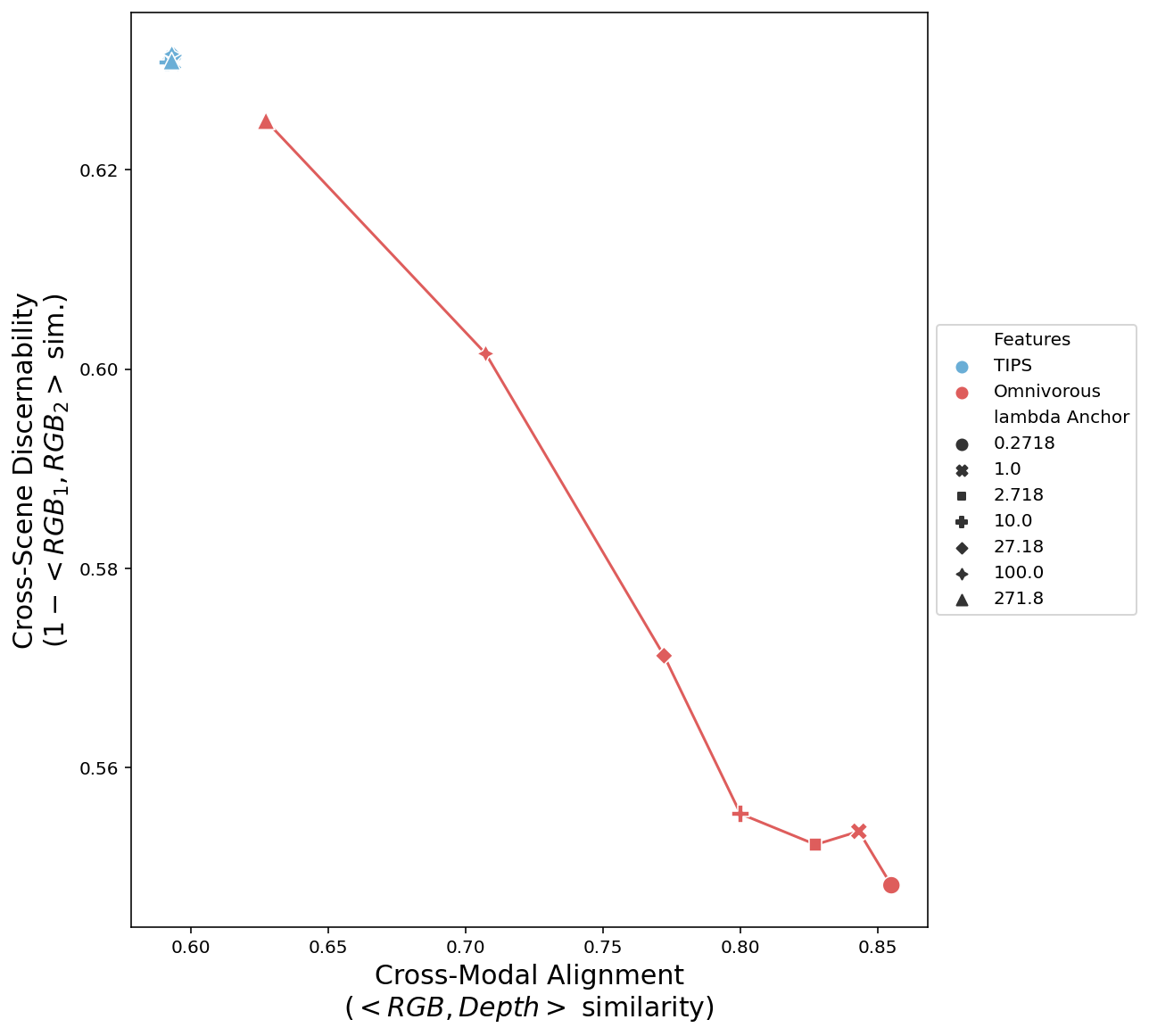}
    \caption{Performance frontier for Omnivorous TIPS (Alignment vs. Discernibility) on TartanAir. We omit the datapoint  for $\lambda_{anchor} = 0.0$, located at $(x=0.783, y=0.859)$, for clarity.	\vspace{-2mm}}
    \label{fig:frontier-plot-tips-diagnostic_metrics}
  \end{subfigure}
  
  \vspace{1em} 

  \begin{subfigure}[b]{0.9\linewidth}
    \centering
    \includegraphics[width=\linewidth]{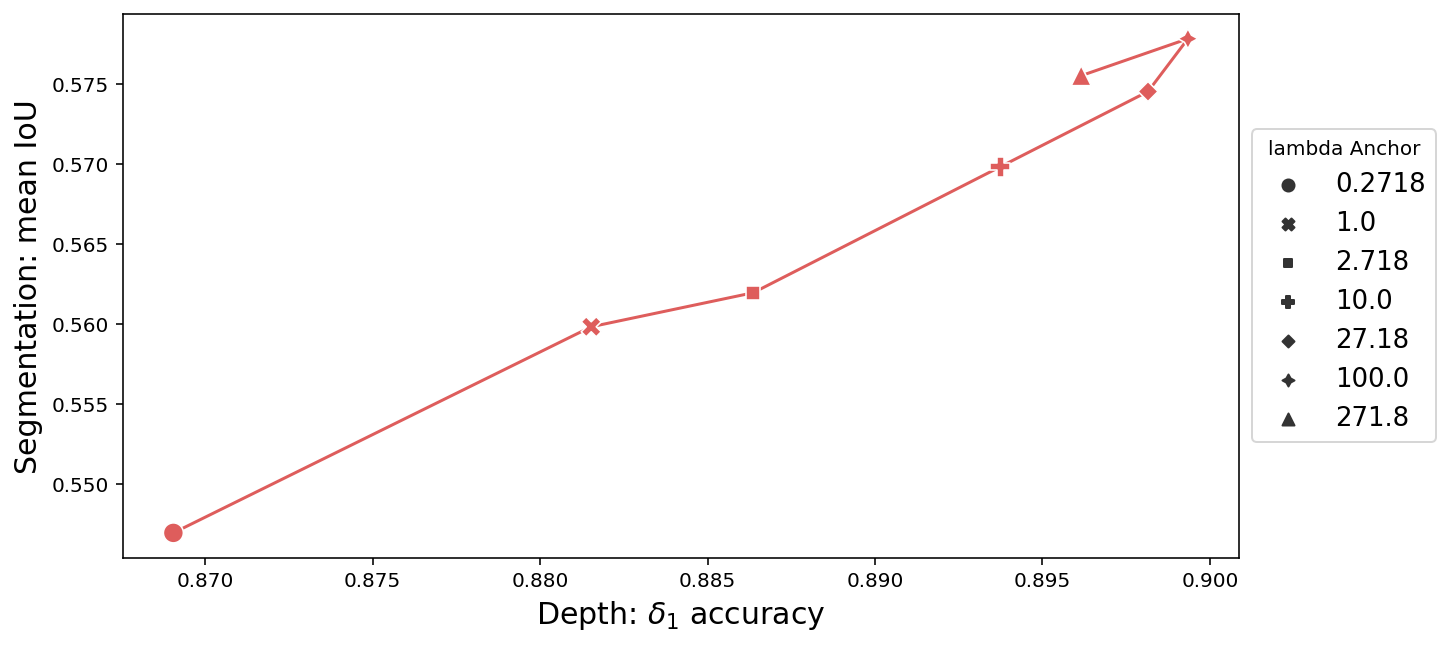} 
    \caption{Performance frontier for Omnivorous TIPS (Segmentation vs. Depth). As in Fig \ref{fig:frontier_plot-segmentation_vs_depth}, we use linear-head evaluation prediction performance for Depth (NYUv2) and Segmentation (Cityscapes). We omit the datapoint for $\lambda_{anchor} = 0.0$, located at $(x=0.745, y=0.435)$, for clarity.}
    \label{fig:frontier_plot-tips-segmentation_vs_depth}
  \end{subfigure}

  \caption{
    \textbf{Behavior of Omnivorous TIPS.}
  }
  \label{fig:frontier_plots-tips}
\end{figure}

Fig \ref{fig:frontier_plots-tips} shows that although it is harder for Omnivorous distillation to improve on the performance of TIPS (than in the case of DINOv2), $\lambda_{anchor}=100$ nevertheless does exceed the depth and segmentation performance of higher values of $\lambda_{anchor}$, which are anchored more strongly to the pretrained teacher. %
This attests to the generality of the Omnivorous framework regardless of the choice of pretrained teacher network.

\subsubsection{Training an Adapter on Top vs. Fine-Tuning Final Blocks}
\label{app:ablation_tips}

We now ablate the parametrization of the student network. Rather than the default setting of fine-tuning the final blocks of a pretrained backbone, we train a zero-initialized adapter network on top of the frozen backbone. In this scenario, the student network is in fact larger than the teacher. All the teacher blocks are frozen and preserved in the student network; only the adapter blocks are trained. We use the same number of adapter blocks (four) as we fine-tuned for our default version of Omnivorous DINOv2.

We evaluate each scenario using a linear head on the final layer. We do not use a DPT head as it would require intermediate activations (typically from blocks [3, 6, 9, 12] in a 12-block ViT-B network), which cannot be consistently applied between the ``adapter-on-top'' and ``finetune-final-blocks'' settings, because the former in fact comprises 16 blocks rather than 12.

Table \ref{tab:ablation_head_or_distillation} shows comparable performance between the two settings, showing our distillation-based approach and training losses can easily be applied to alternative parametrizations of the student network.

\begin{table*}[]
\caption{\textbf{Ablating the parametrization of the student:} we either train a 4-block ViT on top of the DINOv2 ViT-B/14 backbone, or fine-tune the final 4 blocks of the backbone (ours). As before in Table \ref{tab:mixup_ablation}, we report metrics (all $\uparrow$) on (i) classification, using either linear probes on TOK \& GAP, or k-NN, (ii) depth prediction (linear head), (iii) segmentation (linear head), and (iv) multiview correspondence.}
\label{tab:ablation_head_or_distillation}
\begin{tabular}{lcc|cc|ccc|c}
        \toprule
             & \multicolumn{2}{c|}{Classification (acc.)} & \multicolumn{2}{c|}{Depth ($\delta_1$)}       & \multicolumn{3}{c|}{Segmentation (mean IoU)}             & Corresp. (PCK)       \\
dataset             & inet (linear)    & inet (k-NN)     & navi   & nyuv2           & ade20k         & cityscapes     & pascal voc & navi \\ 
parametrization & &  & &  & &  & & \\ \midrule
Adapter on top     & \textbf{0.840} & 81.832          & 0.679          & \textbf{0.905} & 0.470          & 0.628          & 0.826      & 28.15          \\
Fine-tune final blocks & 0.838          & \textbf{81.974} & \textbf{0.706} & 0.896          & \textbf{0.475} & \textbf{0.632} & 0.826      & \textbf{29.00} \\ \bottomrule
\end{tabular}
\end{table*}

\subsubsection{Number of Blocks to Freeze}
\label{app:ablation_layers_to_freeze}

We assess how many ViT blocks can be inherited from the teacher network and kept frozen in Table \ref{tab:ablation_blocks_frozen}. As before, we fine-tune only the final blocks of the network, keeping the preceding $L_{\text{stop-gradient}}$ blocks frozen. We evaluate depth and segmentation prediction using both a DPT and linear head. Our default setting for Omnivorous ViT-B/14, $L_{\text{stop-gradient}}=8$ is chosen on this basis.

\begin{table*}
    \centering
    \caption{\textbf{Ablating the number of blocks kept frozen}, denoted by $L_{\text{stop-gradient}}$, when training Omnivorous DINOv2. There are 12 total blocks in the ViT-B/14 architecture.}
    \label{tab:ablation_blocks_frozen}
        \begin{tabular}{llcc|ccc}
        \toprule
         &  & \multicolumn{2}{c|}{Depth ($\delta_1$)} & \multicolumn{3}{c}{Segmentation (mean IoU)} \\
         & dataset & navi & nyuv2 & ade20k & cityscapes & pascal voc \\
        readout & $L_{\text{stop-gradient}}$ &  &  &  &  &  \\
        \midrule
        \multirow[t]{3}{*}{DPT} & 4 & 0.777 & 0.948 & 0.495 & 0.727 & 0.855 \\
         & 6 & 0.778 & 0.947 & 0.494 & \textbf{0.733} & 0.853 \\
         & 8 & \textbf{0.781} & 0.948 & \textbf{0.505} & 0.732 & \textbf{0.857} \\
         & 10 & 0.780 & \textbf{0.949} & 0.504 & 0.731 & 0.852 \\
        \cline{1-7}
        \multirow[t]{3}{*}{Linear} & 4 & 0.698 & 0.894 & 0.475 & 0.622 & 0.829 \\
         & 6 & 0.703 & 0.896 & \textbf{0.476} & 0.629 & 0.829 \\
         & 8 & \textbf{0.706} & 0.896 & 0.475 & \textbf{0.632} & 0.826 \\
         & 10 & 0.705 & 0.895 & 0.473 & 0.628 & 0.825 \\
        \cline{1-7}
        \bottomrule
        \end{tabular}
\end{table*}

\end{document}